%% file: main.tex
\documentclass[10pt,twocolumn,letterpaper]{article}

\usepackage[pagenumbers]{iccv} 


\definecolor{iccvblue}{rgb}{0.21,0.49,0.74}
\usepackage[pagebackref,breaklinks,colorlinks,allcolors=iccvblue]{hyperref}

\title{Loss Functions for Predictor-based Neural Architecture Search}

\author{Han Ji \quad Yuqi Feng \quad Jiahao Fan \quad Yanan Sun\\
College of Computer Science, Sichuan University\\
{\tt\small jihan@stu.scu.edu.cn, feng770623@gmail.com, \{fanjh,ysun\}@scu.edu.cn}
}

\begin{document}
\maketitle
\input{sec/0_abstract}

\input{sec/1_intro}
\input{sec/2_related}
\input{sec/3_loss}

\input{sec/4_experiment}
\input{sec/5_suggestions}

\input{sec/6_conclusion}

{
    \small
    \bibliographystyle{ieeenat_fullname}
    \bibliography{main}
}
\end{document}

%% file: sec/0_abstract.tex
\begin{abstract}
Evaluation is a critical but costly procedure in neural architecture search (NAS). Performance predictors have been widely adopted to reduce evaluation costs by directly estimating architecture performance. The effectiveness of predictors is heavily influenced by the choice of loss functions. While traditional predictors employ regression loss functions to evaluate the absolute accuracy of architectures, recent approaches have explored various ranking-based loss functions, such as pairwise and listwise ranking losses, to focus on the ranking of architecture performance. Despite their success in NAS, the effectiveness and characteristics of these loss functions have not been thoroughly investigated. In this paper, we conduct the first comprehensive study on loss functions in performance predictors, categorizing them into three main types: regression, ranking, and weighted loss functions. Specifically, we assess eight loss functions using a range of NAS-relevant metrics on 13 tasks across five search spaces. Our results reveal that specific categories of loss functions can be effectively combined to enhance predictor-based NAS. Furthermore, our findings could provide practical guidance for selecting appropriate loss functions for various tasks. We hope this work provides meaningful insights to guide the development of loss functions for predictor-based methods in the NAS community.
\end{abstract}

%% file: sec/1_intro.tex
\section{Introduction}
\label{sec:intro}
Neural architecture search (NAS) aims to automate the design of specialized neural architectures for any given task and has been widely applied in various domains~\cite{liu2018progressive,ghiasi2019fpn,wang2020textnas}. NAS typically contains three key components: search space, search strategy, and performance evaluation~\cite{elsken2019neural}. One significant bottleneck of early NAS was the expensive performance evaluation stage, which involves training each architecture from scratch~\cite{zoph2016neural,real2017large}. To mitigate the heavy computational cost, prior research has introduced a wide range of evaluation acceleration methods, such as weight sharing~\cite{pham2018efficient}, early stopping strategy~\cite{falkner2018bohb}, zero-cost proxy~\cite{abdelfattah2021zero}, and performance predictor~\cite{liu2021homogeneous}.

Among these methods, performance predictors are one of the most widely used because of their ability to provide accurate and stable predictions across various search spaces. A performance predictor requires a small number of trained architectures to learn the mapping from the encoding of an architecture to its corresponding performance. After that, it can directly estimate the performance of unseen architectures, significantly reducing the need for expensive evaluations.  

A crucial component of a performance predictor is the loss function, which determines its optimization direction. Previous studies~\cite{ning2020generic,xu2021renas} have demonstrated that the choice of loss function can dramatically affect the performance of the predictor. This raises an important question: \textit {how to design a proper loss function to empower the performance predictor?} A large number of works~\cite{wen2020neural,wei2022npenas,lu2023pinat} opt for Mean Square Error (MSE) loss to estimate the absolute accuracy of an architecture. Other works have explored ranking loss functions, such as hinge ranking (HR) loss~\cite{ning2020generic,hwang2024flowerformer}, which predict rankings of architecture performance. Additionally, some works~\cite{shi2020bridging,white2021bananas} employ weighted loss functions, which focus more on architectures with higher accuracy. While several works~\cite{ning2020generic,xu2021renas} have discussed the effect of loss functions in predictors, they only consider MSE and ranking loss. Additionally, existing evaluations of loss functions often lack generalizability due to limited search spaces and assessment metrics. These limitations motivate us to investigate how different loss functions compare across various settings and how their respective strengths can be leveraged to enhance predictor-based NAS.

\input{figs/pipeline.tex}

In this paper, we provide the first comprehensive study on loss functions in performance predictors. We identify three significant categories of loss functions. \textbf{Regression} loss functions minimize the difference between the prediction score and the absolute accuracy of an architecture. \textbf{Ranking} loss functions aim to rank architectures correctly. They can be further classified into pairwise ones that reduce the number of misranked architecture pairs and listwise ones that align the predicted ranking list with the ground truth of architectures. \textbf{Weighted} loss functions assign weights to an architecture according to its absolute accuracy. The main categories of loss functions are illustrated in Figure~\ref{fig:pipeline}.

We conduct rigorous experiments on 13 tasks across five popular search spaces: NAS-Bench-101~\cite{ying2019bench}, NAS-Bench-201~\cite{dong2019bench}, TransNAS-Bench-101 Micro/Macro~\cite{duan2021transnas}, NAS-Bench-Graph~\cite{qin2022bench}, and DARTS~\cite{liu2018darts}. Each loss function is evaluated based on two aspects: overall ranking performance and performance on identifying well-performing architectures. The latter is particularly important because the primary goal of NAS is to discover the best architectures within a search space. Our experiments are performed under various settings, including different numbers of training data and types of predictors. Finally, we evaluate the performance of each loss function in predictor-based NAS.

Overall, our results reveal that loss functions have a substantial impact on the effectiveness of predictors, especially in identifying well-performing architectures. Besides, we also find that no loss functions have consistent performance across different search spaces, numbers of training data, and types of predictors. In general, weighted loss functions perform better at distinguishing good architectures with enough training data, but they can be inferior to ranking loss functions when training data are extremely limited. Furthermore, we demonstrate that combining effective loss functions into a piecewise (PW) loss function can lead to significant performance improvements for predictor-based NAS. Our comprehensive findings unveil the crucial role of loss functions in performance predictors. We hope this study will offer valuable insights for the design of new loss functions for predictor-based NAS.

\noindent\textbf{Our contributions} are summarized below:
\begin{itemize}
    \item We provide the first comprehensive study on a series of loss functions in performance predictors, including regression, ranking, and weighted loss functions.
   
    \item We conduct extensive experiments on the effectiveness of loss functions on five search spaces. Our results can offer recommendations for the optimal loss function to use for different NAS tasks.

    \item We demonstrate that specific categories of loss functions can be combined to enhance the predictor-based NAS framework, leading to performance improvements over any single loss as well as previous competitive methods.

\end{itemize}

%% file: figs/pipeline.tex
\begin{figure*}[t]
        \centering
        \includegraphics[width=0.95\textwidth]{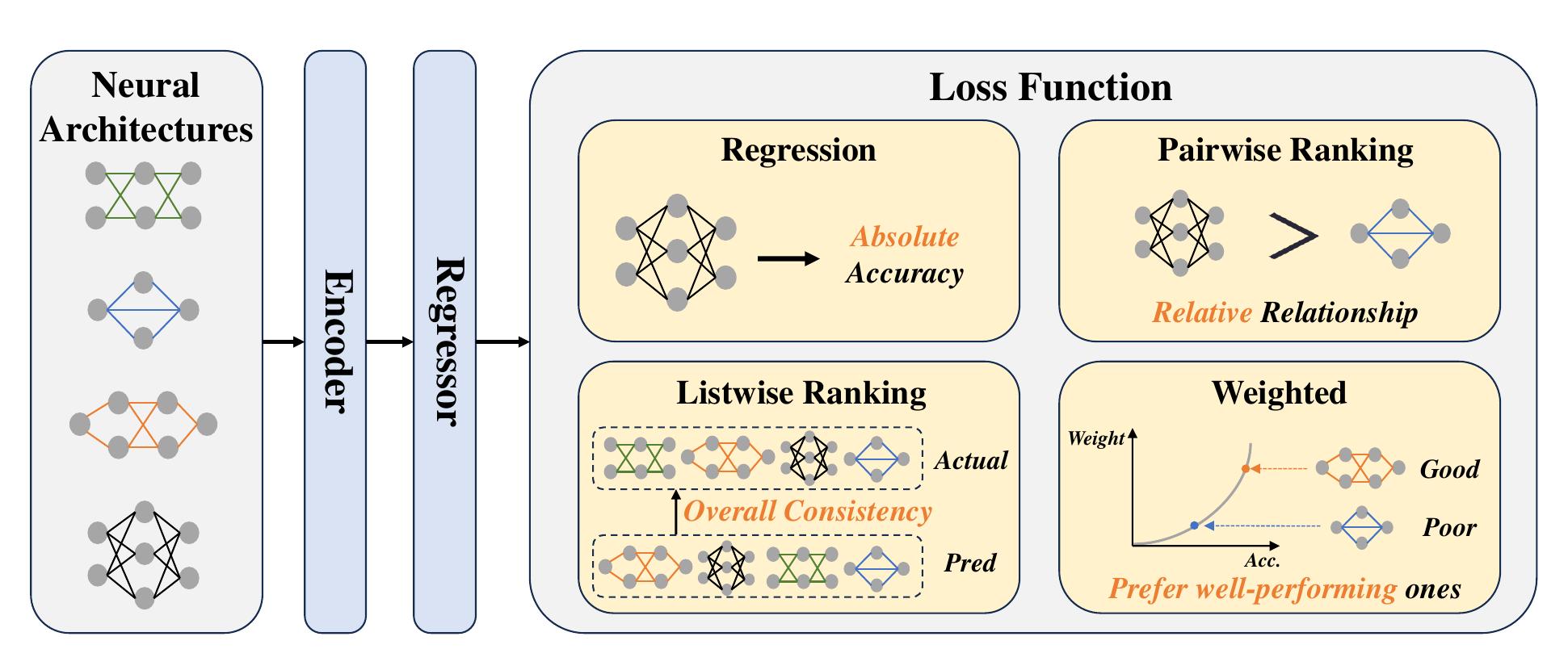}
     \caption{Illustration of main loss functions discussed in this paper. Ranking loss functions are further classified into pairwise and listwise ones. During the training stage of a predictor, it first obtains the architecture representation from the encoder and then regresses representation to the prediction score via a loss function. The choice of loss function decides the optimization direction of the predictor.}
    \label{fig:pipeline}
\end{figure*}

%% file: sec/2_related.tex
\section{Relate Works}
\label{sec:related}

\noindent\textbf{Performance Predictors for NAS}. Performance predictors can estimate the final performance of unseen architectures after learning from a limited set of trained architectures, enabling NAS to explore the search space efficiently. The effectiveness of predictors largely depends on the quantity of training data, which is typically obtained by training architectures from scratch. To construct an effective predictor with fewer training samples, recent studies mainly focus on generating informative architecture representation by designing advanced models~\cite{wen2020neural,lu2023pinat}, optimizing architecture encoding~\cite{ning2020generic,yan2021cate}, and incorporating tailored self-supervised tasks~\cite{jing2022graph,zheng2024dclp}. Despite these advancements, the impact of loss functions on performance predictors remains underexplored.

\noindent\textbf{Loss Functions for Performance Predictors}. Several loss functions have been employed in performance predictors. Early predictors~\cite{luo2018neural,liu2021homogeneous,lu2021tnasp} typically employ regression loss functions such as MSE to predict architecture accuracy directly. Another widely adopted paradigm is ranking-based loss functions, which focus on learning the relative ordering of architectures. For instance, GATES~\cite{ning2020generic}, Arch-Graph~\cite{huang2022arch}, and GMAENAS~\cite{jing2022graph} separately utilize pairwise HR, Binary Cross Entropy (BCE), and Bayesian Personalized Ranking (BPR) loss to estimate the relative rankings of architectures. Additionally, DCLP~\cite{zheng2024dclp} uses ListMLE~\cite{xia2008listwise}, a listwise ranking loss function, to generate a ranking list that closely aligns with ground truth rankings. Hybrid loss functions have also been explored to balance accuracy prediction and ranking prediction. For example, NAR-Former~\cite{yi2023nar} combines MSE with pairwise Sequence Ranking (SR) loss to simultaneously preserve the absolute accuracy of each architecture while refining relative rankings. Finally, weighted loss functions, such as Exponential Weighted (EW)~\cite{shi2020bridging} and Mean Absolute Percentage Error (MAPE)~\cite{white2021bananas} loss, assign higher weights to well-performing architectures, prioritizing predictive precision for high-accuracy candidates. These weighted losses are typically derived from regression loss functions. Although some works employ auxiliary loss functions like unsupervised learning loss~\cite{yan2021cate} and consistency loss~\cite{yi2023nar} to enhance the performance predictor, they fall beyond the scope of our study. This paper focuses on analyzing the core loss functions used in performance predictors.

%% file: sec/3_loss.tex
\section{Loss Functions of Performance Predictors}
We denote a set of architectures sampled from search space $S$ as $X=\{x_1,x_2,\dots,x_n\}$ and their respective ground-truth (GT) performances as $Y=\{y_1,y_2,\dots,y_n\}$. Then we have a training set $D=\{(x_1,y_1),(x_2,y_2),\dots,(x_n,y_n)\}$ composed of $n$ architecture-GT pairs. We define a mapping function for the predictor $P: X\rightarrow Y$. $P$ takes architectures $X$ as input and output their prediction scores $\hat{Y}=\{\hat{y_1},\hat{y_2},\dots,\hat{y_n}\}$. In this paper, we discuss eight loss functions, containing seven existing ones and a novel one. We split them into four categories and describe them below. Detailed information about these loss functions can be found in the Supplementary Material.

\noindent\textbf{Regression} loss functions are popular options for predictors. For each input architecture $x_i$, the predictor aims to minimize the difference between prediction scores and GTs. \textit{MSE} is selected as the representative loss~\cite{wen2020neural}.
 
\noindent\textbf{Pairwise Ranking} loss functions are also common for predictors, which focus on the relative relationship of architectures. For each input architecture pair $(x_i,x_j)$, the predictor is penalized when the pair is ranked incorrectly. \textit{HR}~\cite{ning2022ta}, \textit{Logistic Ranking (LR)}~\cite{xu2021renas}, and \textit{MSE+SR}~\cite{yi2023nar} are selected as the representative loss. Note that MSE+SR combines the pairwise SR loss with MSE.

\noindent\textbf{Listwise Ranking} loss functions optimize the consistency between the list of prediction scores and the actual ranking of architectures in a listwise manner. We use \textit{ListMLE} loss~\cite{xia2008listwise} as the representative one.

\input{figs/search_spaces_perf.tex}

\noindent\textbf{Weighted} loss functions give greater weights to architectures with higher accuracy to better identify well-performing architectures. We use \textit{EW}\cite{shi2020bridging} and \textit{MAPE}~\cite{white2021bananas} loss as the representative loss. Besides, we introduce \textit{Weighted Approximate-Rank Pairwise (WARP)}~\cite{weston2011wsabie} loss, a weighted pairwise ranking loss widely used in the multi-label image annotation task, into predictor-based NAS.

%% file: figs/search_spaces_perf.tex
\begin{figure*}[t]
     \begin{minipage}[b]{0.19\textwidth}
        \centering
        \includegraphics[width=\textwidth]{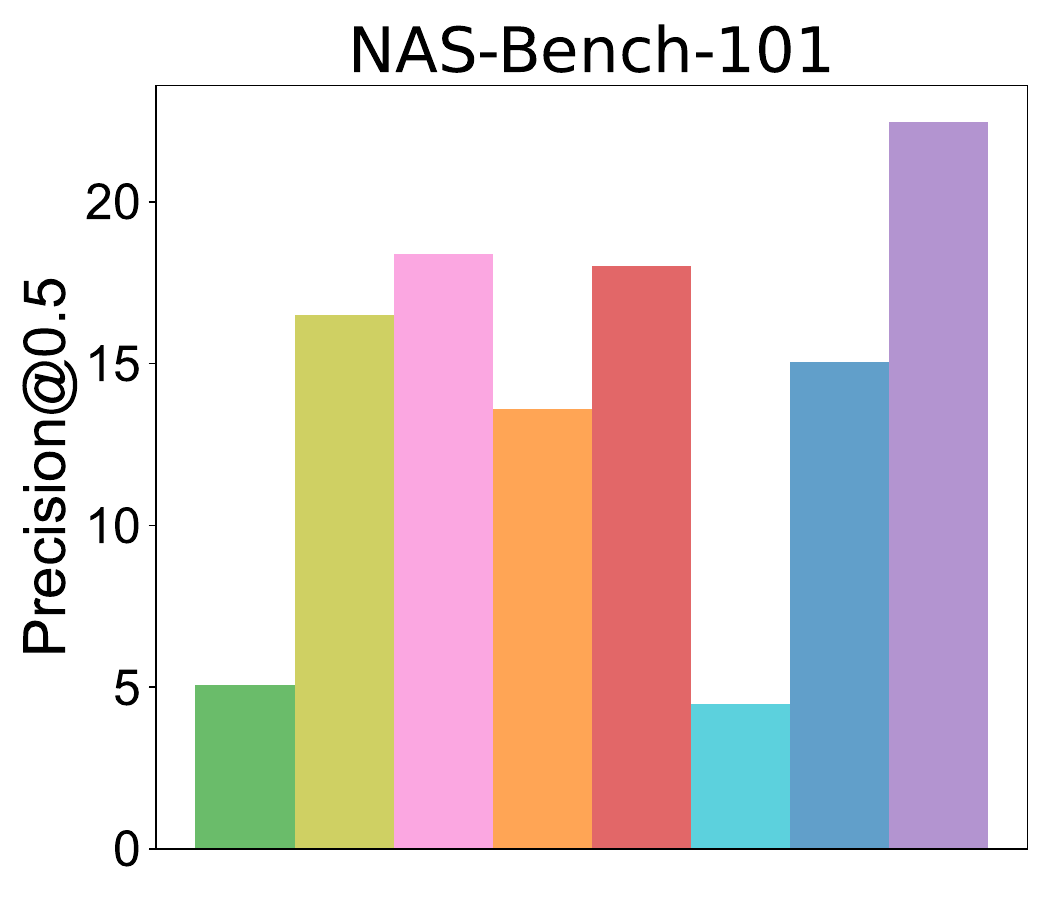}
    \end{minipage}\hfill   
     \begin{minipage}[b]{0.19\textwidth}
        \centering
        \includegraphics[width=\textwidth]{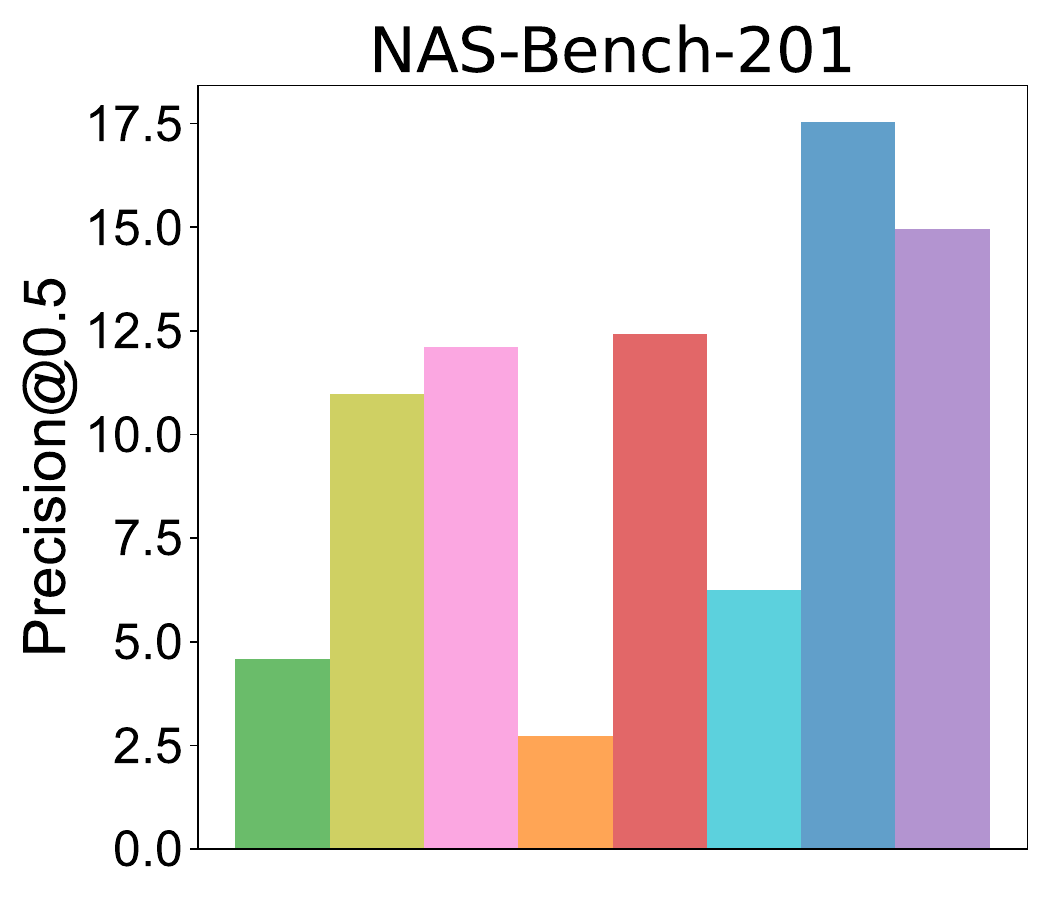}
    \end{minipage}\hfill   
    \begin{minipage}[b]{0.19\textwidth}
        \centering
        \includegraphics[width=\textwidth]{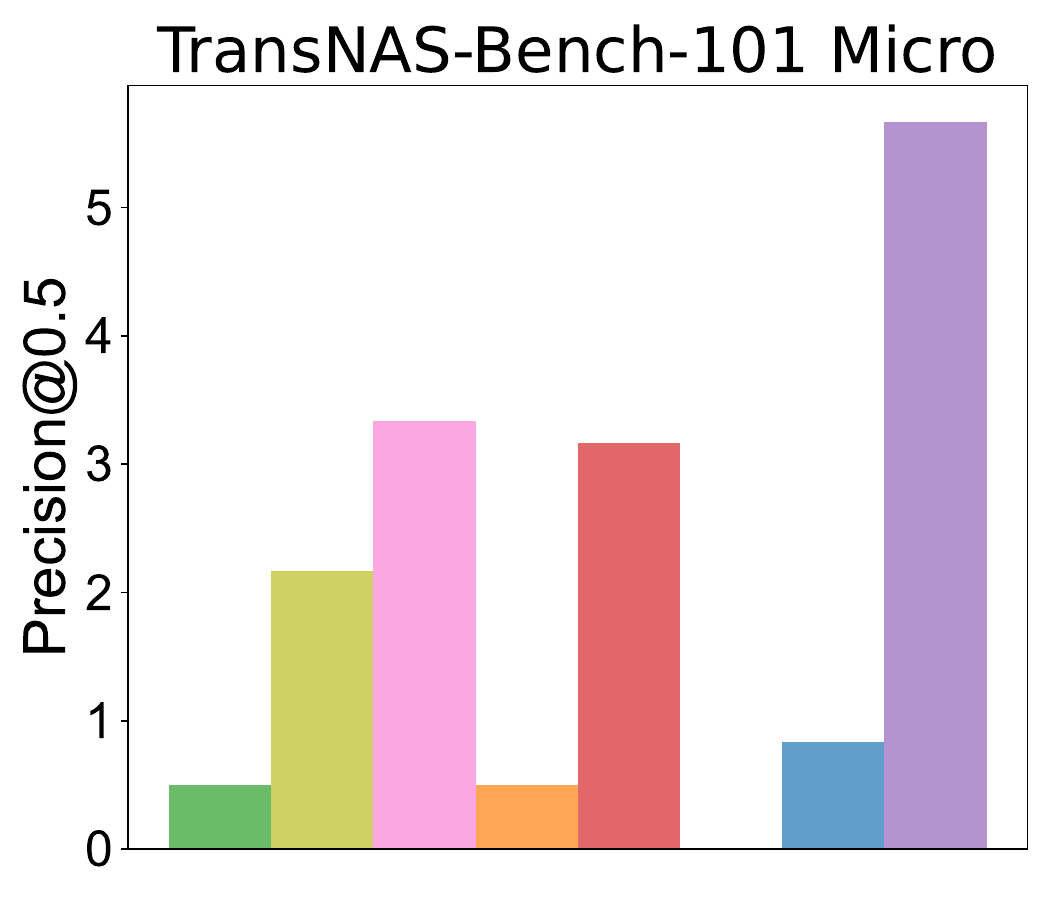}
    \end{minipage}\hfill
      \begin{minipage}[b]{0.19\textwidth}
        \centering
        \includegraphics[width=\textwidth]{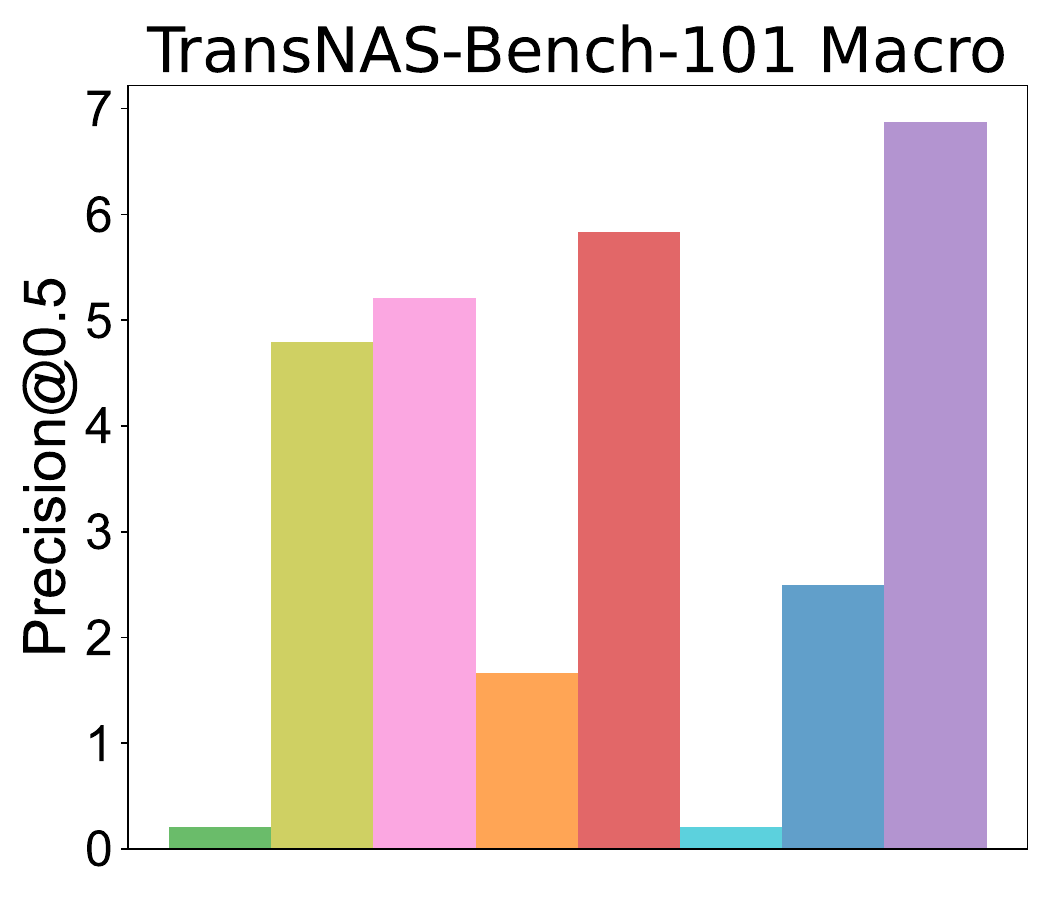}
    \end{minipage}\hfill   
      \begin{minipage}[b]{0.19\textwidth}
        \centering
        \includegraphics[width=\textwidth]{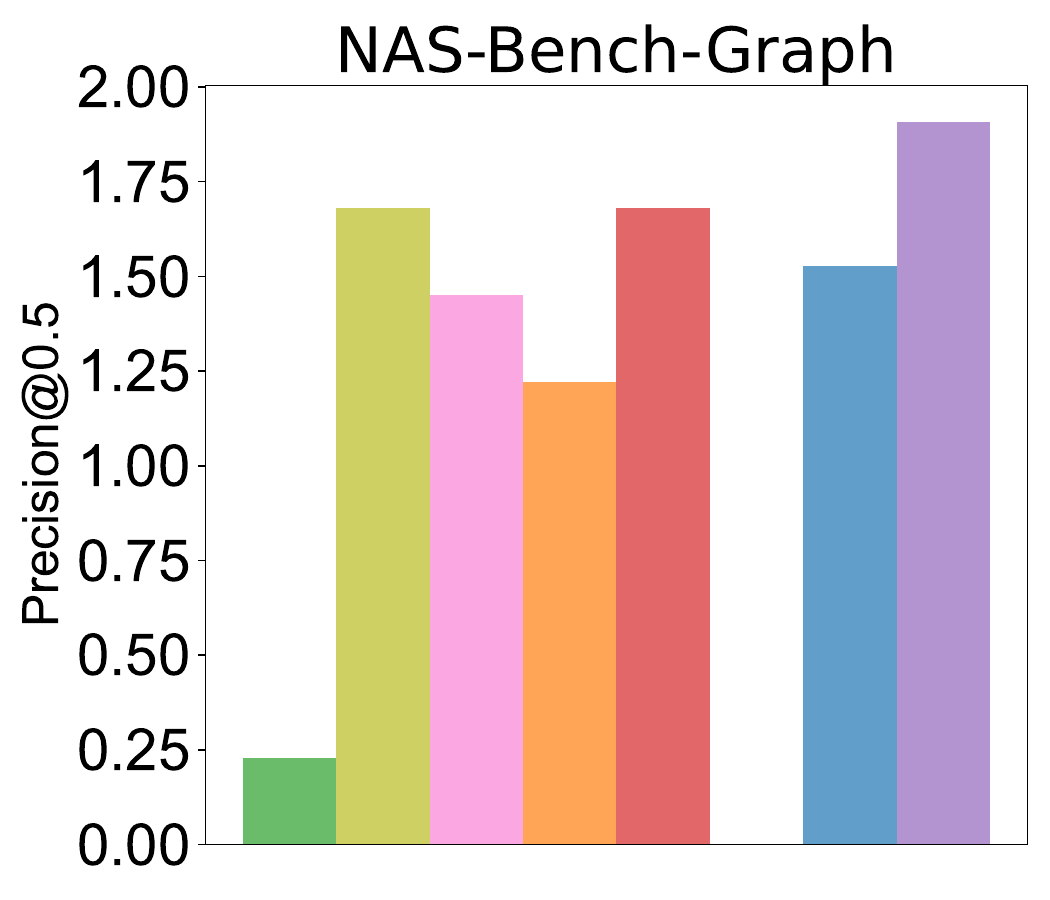}
    \end{minipage}   
    \begin{minipage}[b]{0.19\textwidth}
        \centering
        \includegraphics[width=\textwidth]{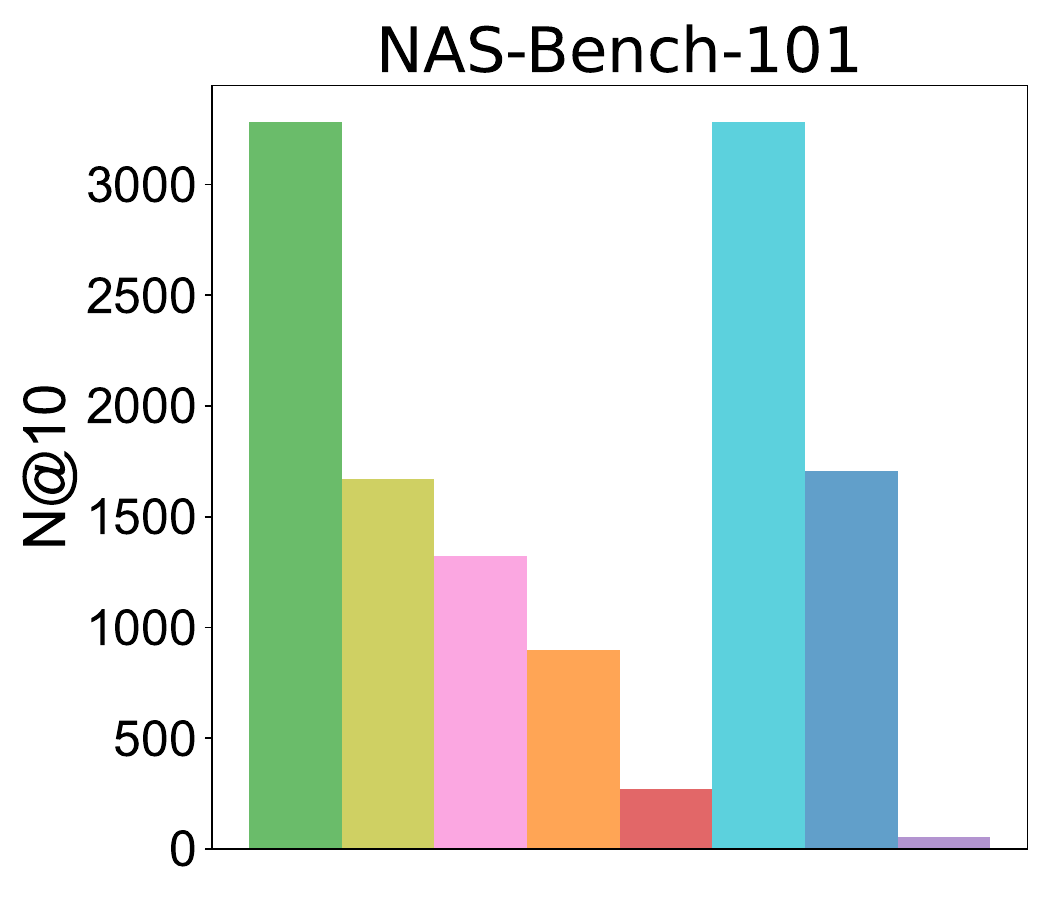}
    \end{minipage}\hfill   
     \begin{minipage}[b]{0.19\textwidth}
        \centering
        \includegraphics[width=\textwidth]{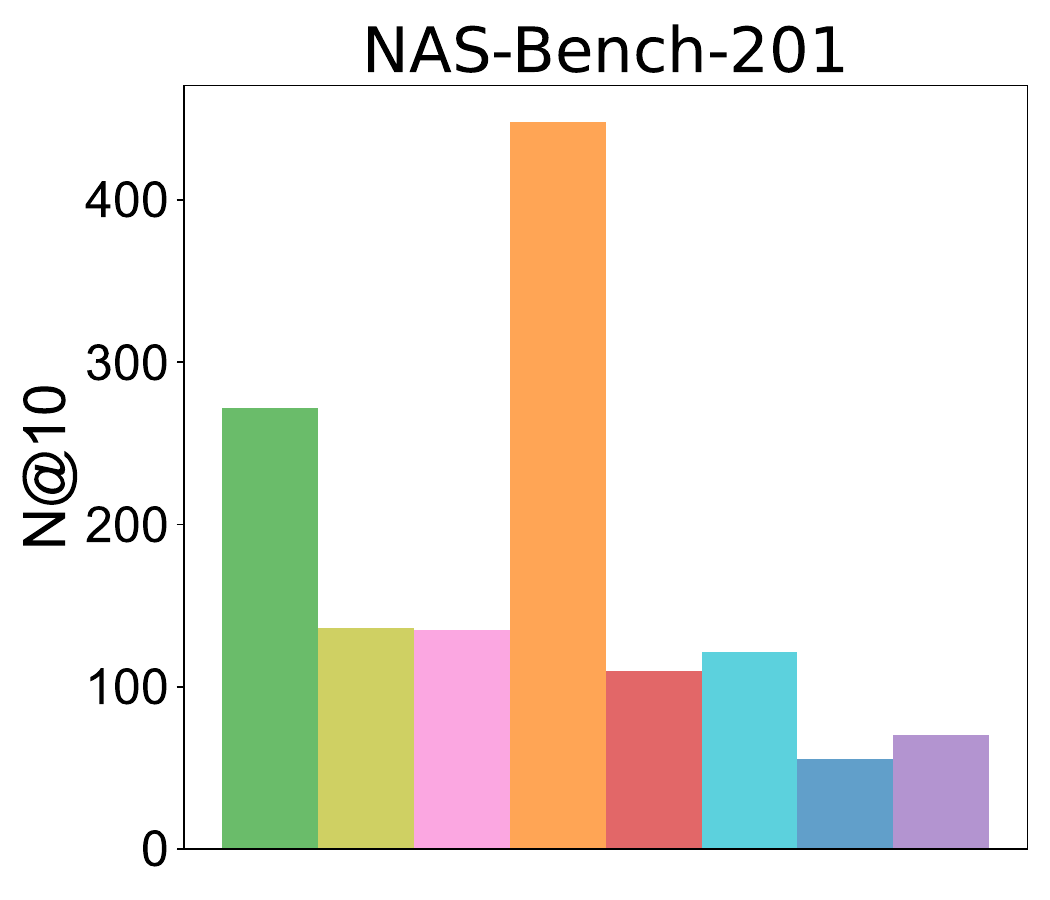}
    \end{minipage}\hfill   
    \begin{minipage}[b]{0.19\textwidth}
        \centering
        \includegraphics[width=\textwidth]{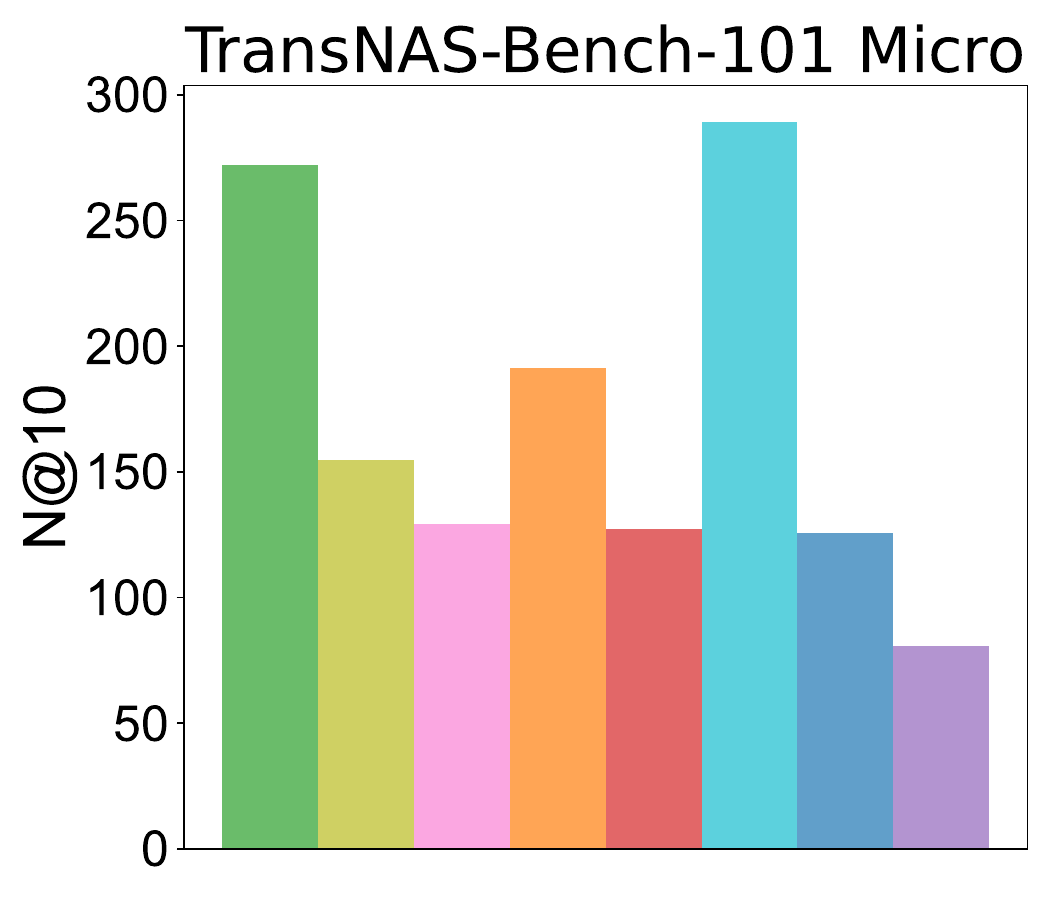}
    \end{minipage}\hfill
      \begin{minipage}[b]{0.19\textwidth}
        \centering
        \includegraphics[width=\textwidth]{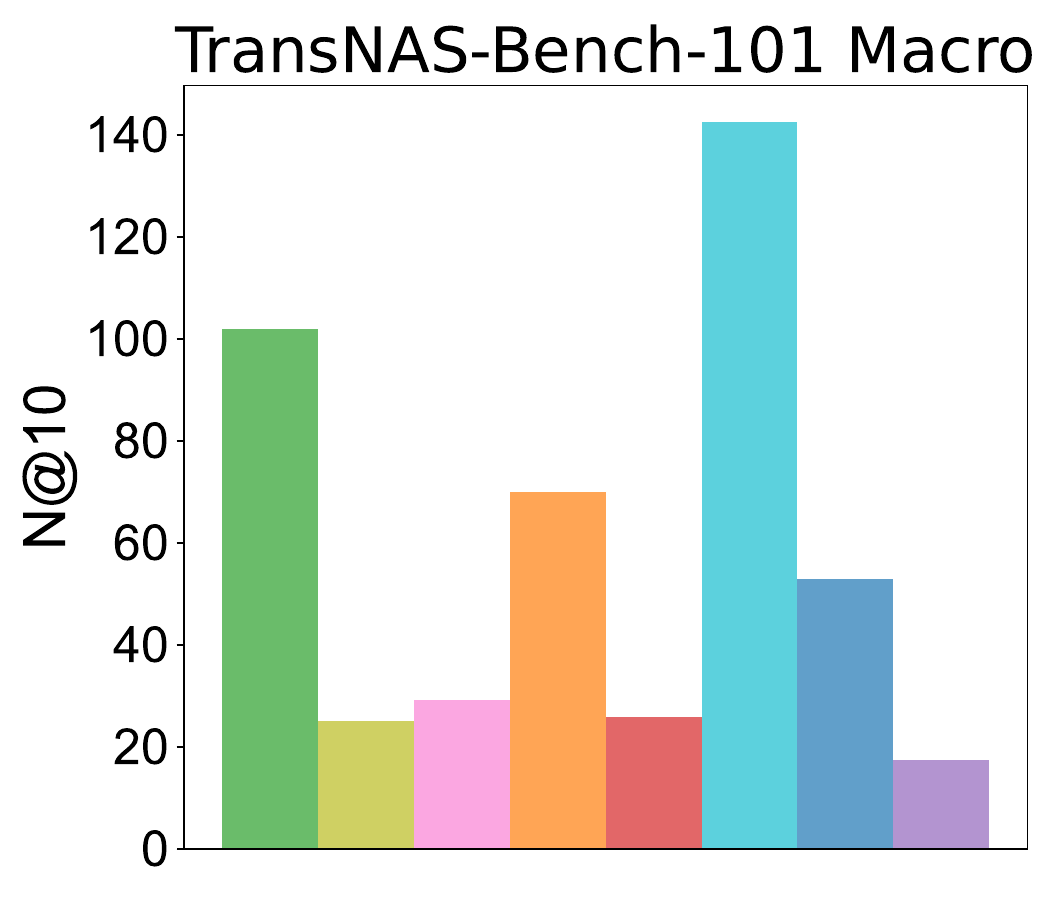}
    \end{minipage}\hfill
      \begin{minipage}[b]{0.19\textwidth}
        \centering
        \includegraphics[width=\textwidth]{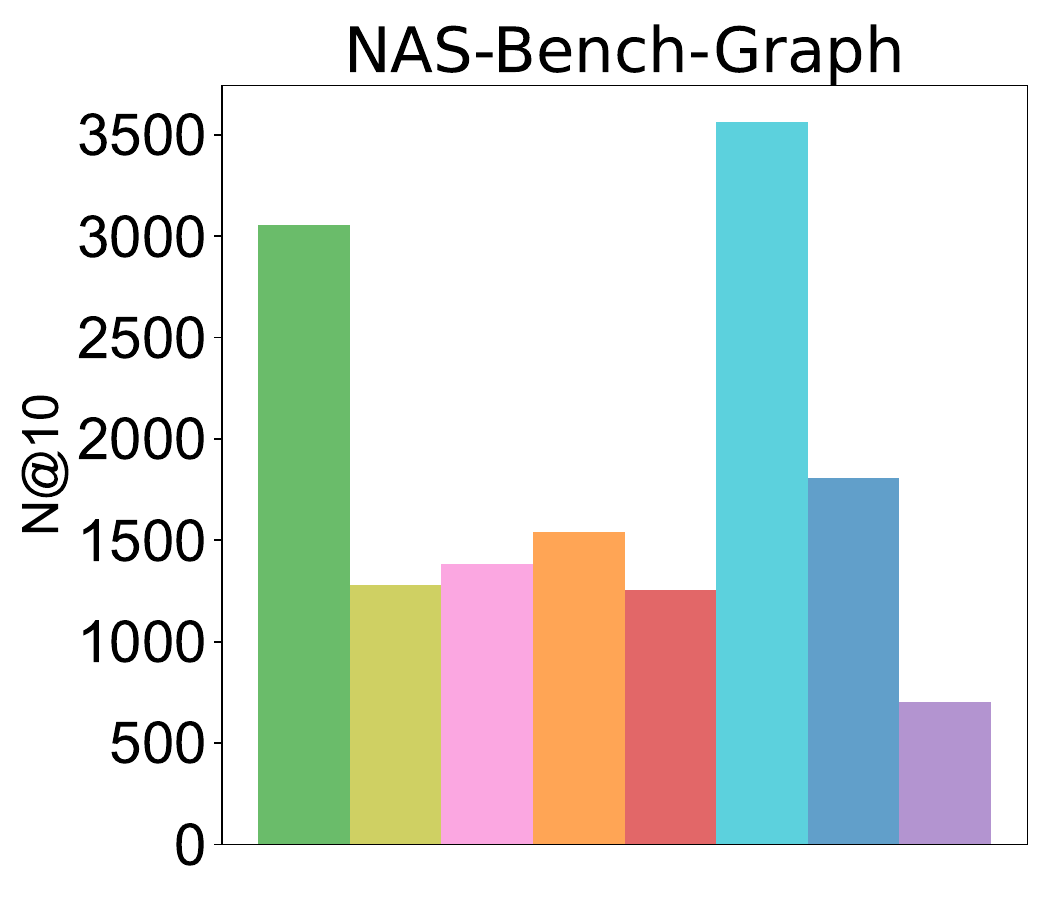}
    \end{minipage}   
    \begin{minipage}[b]{0.19\textwidth}
        \centering
        \includegraphics[width=\textwidth]{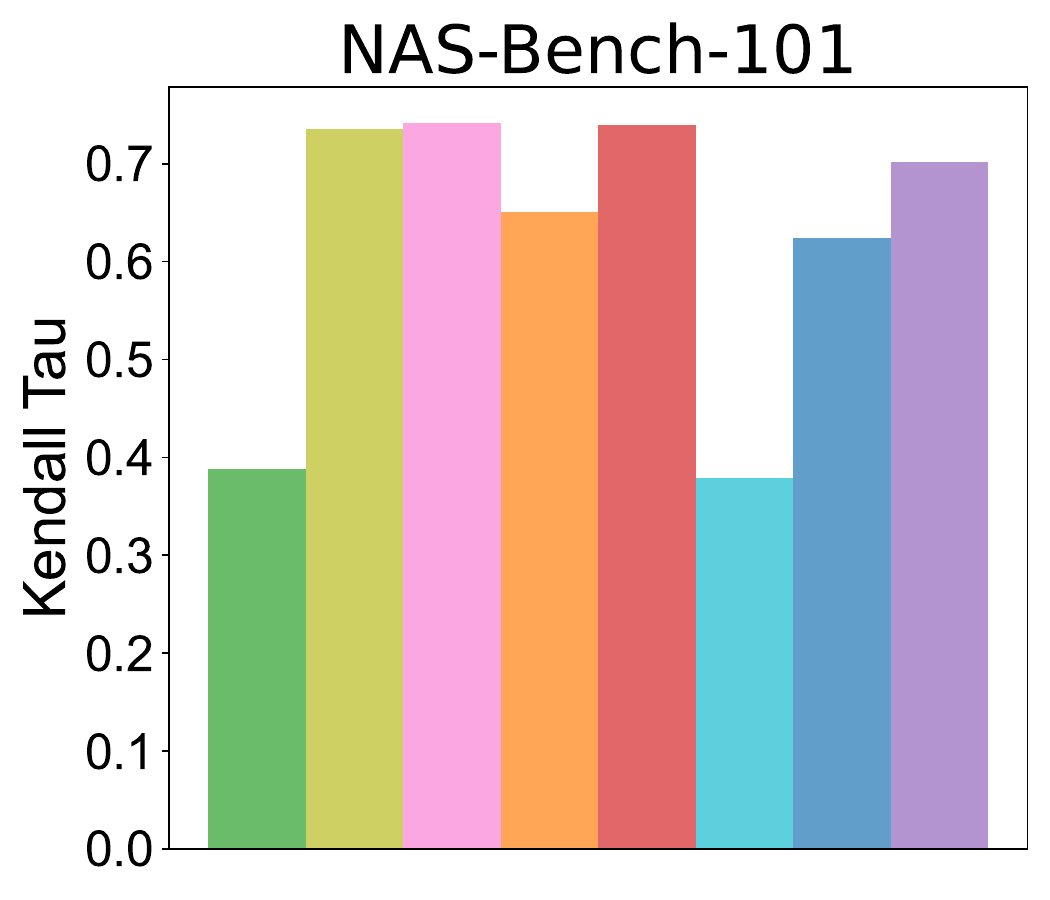}
    \end{minipage}\hfill   
     \begin{minipage}[b]{0.19\textwidth}
        \centering
        \includegraphics[width=\textwidth]{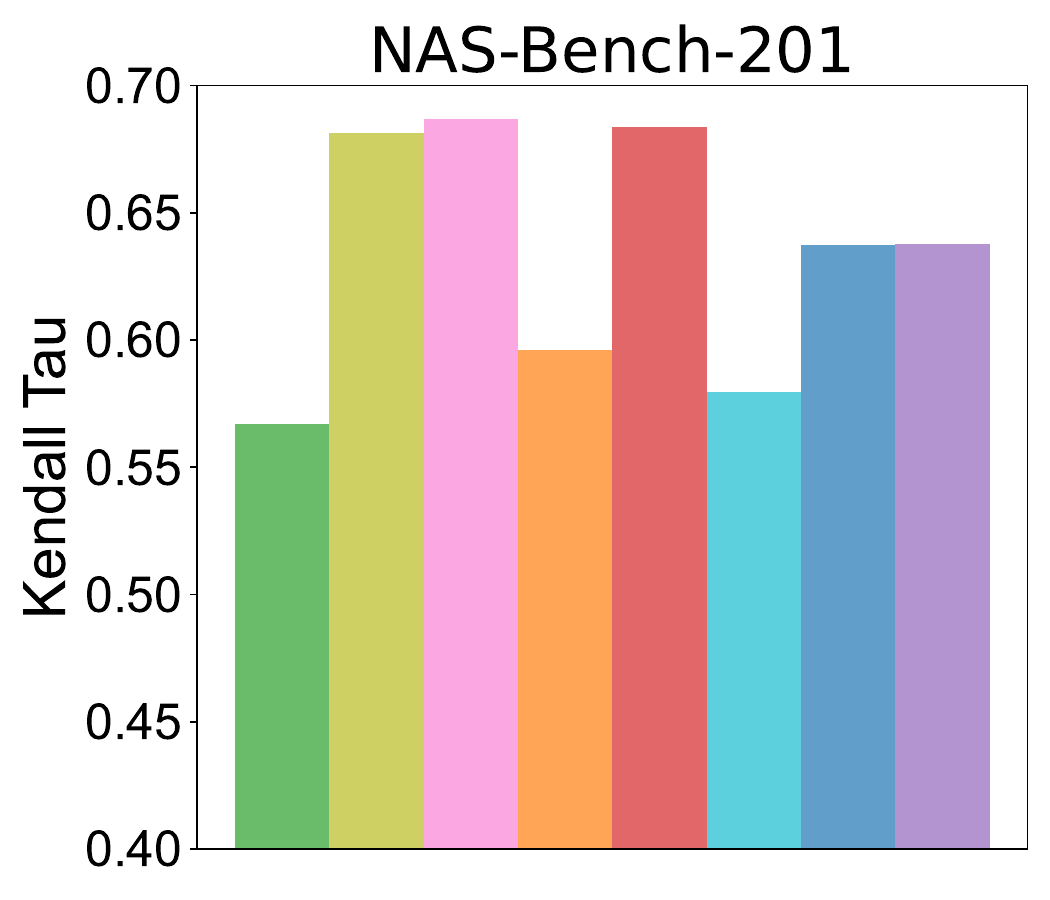}
    \end{minipage}\hfill   
    \begin{minipage}[b]{0.19\textwidth}
        \centering
        \includegraphics[width=\textwidth]{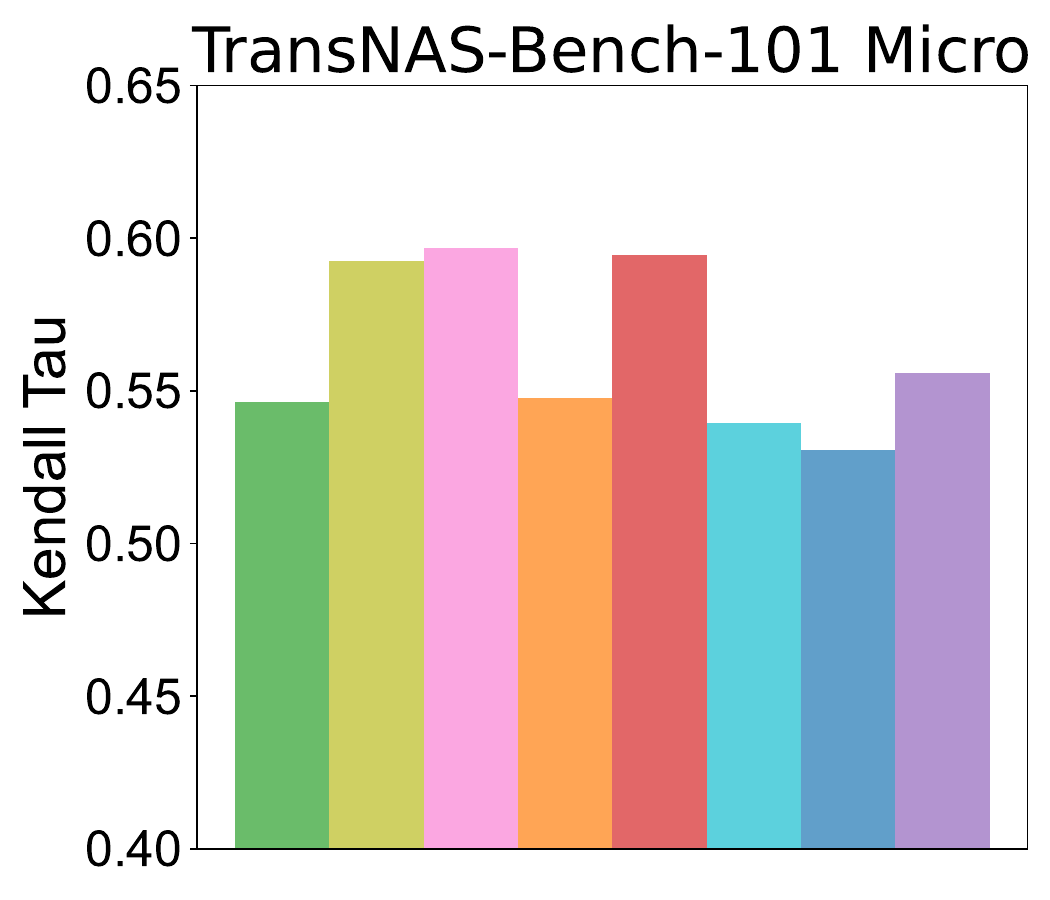}
    \end{minipage}\hfill
      \begin{minipage}[b]{0.19\textwidth}
        \centering
        \includegraphics[width=\textwidth]{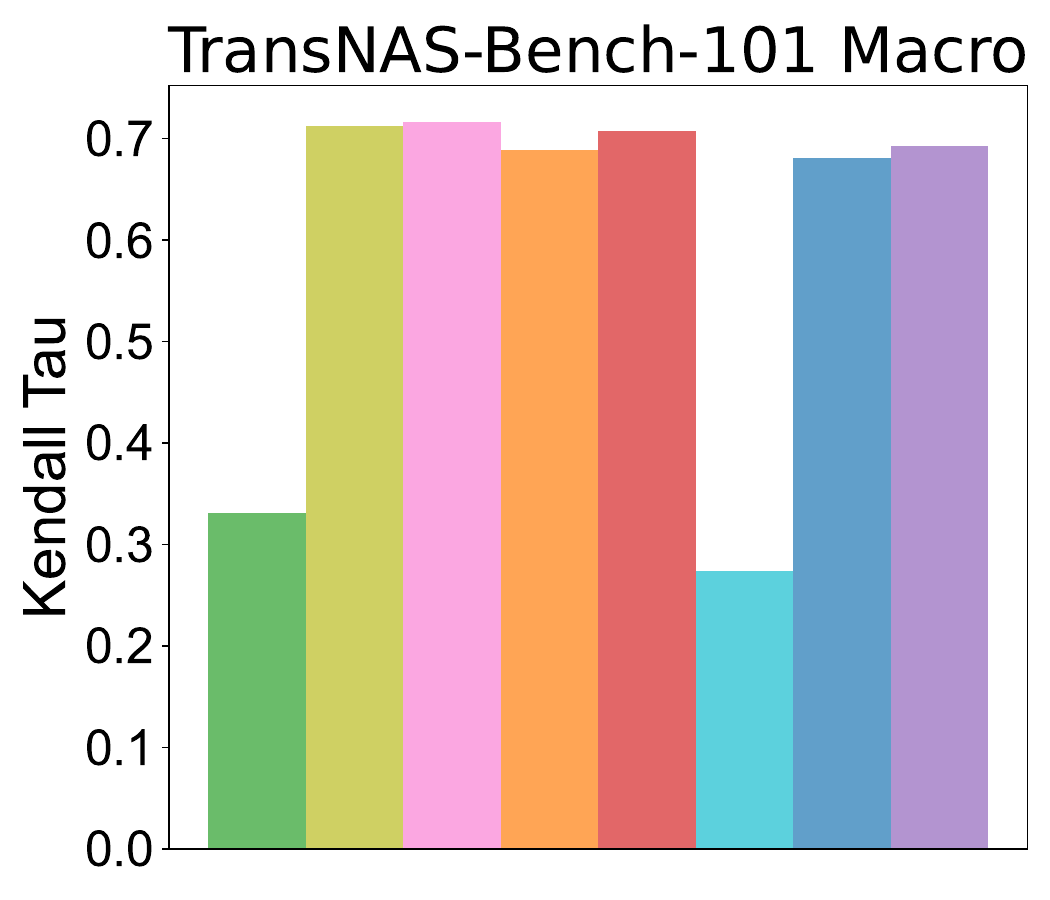}
    \end{minipage}\hfill   
      \begin{minipage}[b]{0.19\textwidth}
        \centering
        \includegraphics[width=\textwidth]{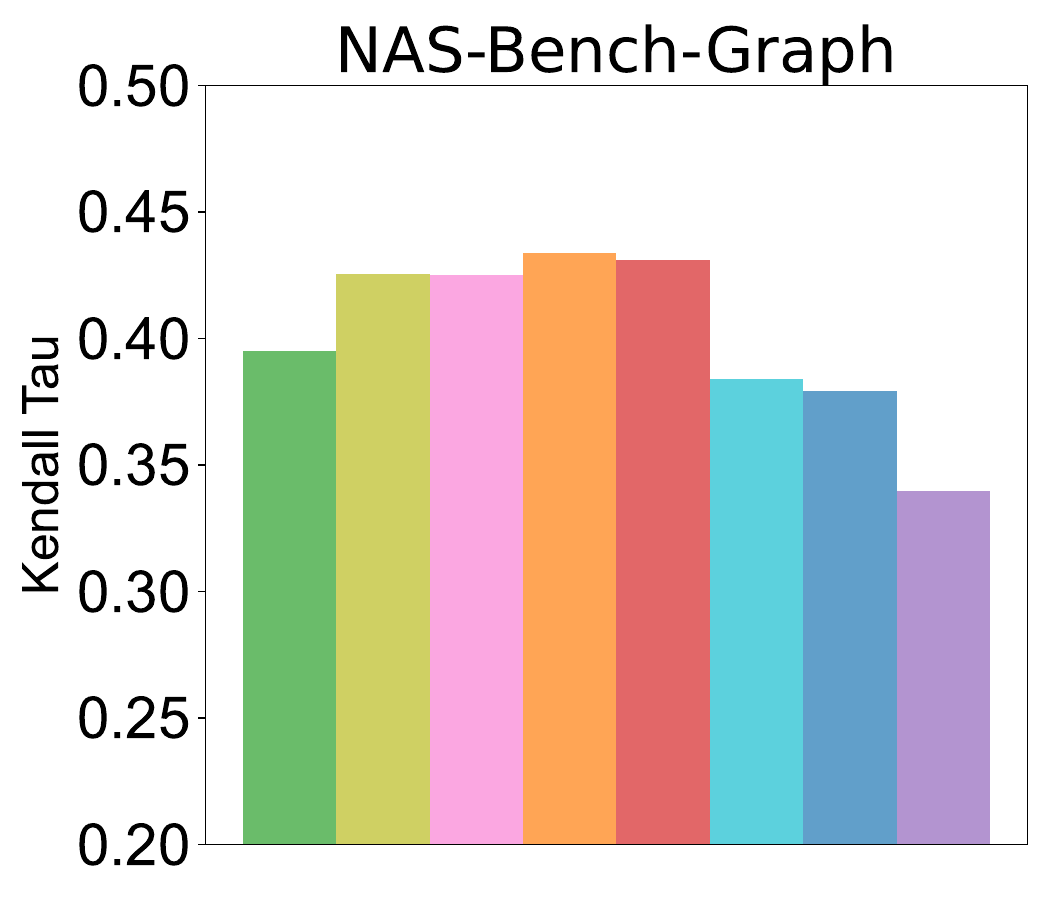}
    \end{minipage}
    \begin{minipage}[b]{0.95\textwidth}
        \centering
         \hspace{20cm}\includegraphics[width=\textwidth]{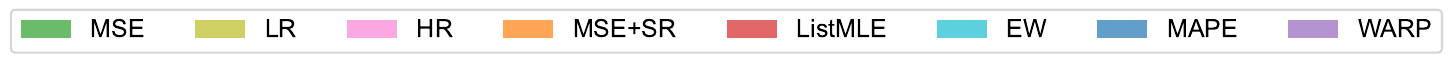}
    \end{minipage}
     \caption{Precision@0.5, N@10, and Kendall's Tau (\textit{Top to Bottom}) of different loss functions. The training portion used on these search spaces is 0.1\%, 1\%, 0.3\%, 0.3\%, and 3\% (\textit{Left to Right}). Note that a higher metric indicates better performance except for N@10.}
    \label{fig:res on search spaces}
\end{figure*}

%% file: sec/4_experiment.tex
\section{Experiments}
\label{sec: evaluation}
In this section, we present the experimental settings and results. Our experiments can be divided into two parts: evaluating the performance of loss functions using a variety of NAS-correlated metrics and evaluating the effectiveness of loss functions in predictor-based NAS. We first discuss the settings used in our experiments.  

\subsection{Experimental Settings}
\noindent\textbf{NAS Search Spaces.} The experiments are conducted in five popular search spaces: NAS-Bench-101~\cite{ying2019bench}, NAS-Bench-201~\cite{dong2019bench}, TransNAS-Bench-101 Micro/Macro~\cite{duan2021transnas}, NAS-Bench-Graph~\cite{qin2022bench}, DARTS~\cite{liu2018darts}. Note that TransNAS-Bench-101 Macro is a skeleton-based search space while the remaining are cell-based ones. 

\noindent\textbf{Assessment Metrics.} We introduce the major metrics to assess loss functions, including the ranking correlation metric and top-ranking precision metrics. We first discuss the former below.

\noindent\textit{Kendall's Tau}~\cite{kendall1938new} ($\tau$) reflects the ordinal association between prediction scores and GTs. It is calculated as $\tau = \frac{2({n_c}-{n_d})}{n(n-1)}$, where $n_c$ and $n_d$ separately denote the number of concordant and discordant architecture pairs.

Since the ability to identify well-performing architectures matters more than distinguishing the bad ones~\cite{ning2021evaluating}, we focus more on the metrics that evaluate the top-K ranking precision. Assuming the actual ranking and the predicted ranking of architecture $x_i$ is $r_i$ and $\hat{r_i}$, respectively. The GT of the $i$-th best architecture in the search space is $y_{g_i}$. We use $X_K=\{x_i|\hat{r_i}<=K\}$ to represent the architectures with top-K highest prediction scores in the search space. $N$ denotes the total number of architectures in the search space. We employ the following two popular metrics~\cite{ning2020generic,hwang2024flowerformer}:

\noindent\textit{Precision@T} $\vspace{2pt} \in (0,100] = \frac{\#\{i|r_i<=TN\cap \hat{r_i}<=TN\}}{TN} $: The portion of actual top-T\% architectures among the top-T\% predicted architectures. Note that $T\in(0,100]$ denotes the portion instead of a specific ranking in this metric.

\noindent\textit{N@K} $ \in [1,N] = \arg\min_{x_i\in X_K} r_i $: The best actual ranking among architectures with the top-K prediction scores. This metric is particularly important as it directly reflects the result of a basic predictor-based NAS paradigm, where a predictor is trained and used to select top-$K$ architectures from candidates. A lower N@K is preferred.

\noindent\textbf{Hyperparameter Tuning.} The effectiveness of each loss function is closely related to the hyperparameter settings. For example, listwise ranking loss functions prefer a larger learning rate. Hence, we use different levels of hyperparameters for each loss function. To achieve a fair comparison, we opt for a uniform Graph Convolutional Network (GCN)-based~\cite{kipf2016semi} performance predictor for its success in prior works~\citep{wen2020neural,dudziak2020brp,shi2020bridging,liu2022bridge}.
The hyperparameters for loss functions are reported in the Supplementary Material. 

\input{figs/training_portion.tex}

\subsection{Performance Predictor Evaluation}
\label{sec: predictor evaluation}
 We conduct this set of experiments on NAS-Bench-101~\cite{ying2019bench}, NAS-Bench-201~\cite{dong2019bench}, TransNAS-Bench-101 Micro/Macro~\cite{duan2021transnas}, NAS-Bench-Graph~\cite{qin2022bench}. For a more consistent evaluation, the predictor is trained on a subset of each search space and evaluated on the whole search space unless stated otherwise. All these search spaces provide both validation and test accuracy for each architecture. Following previous works~\cite{lu2023pinat,lu2021tnasp}, we use the former to train the predictor and evaluate it with the latter. Since several search spaces include multiple tasks, we only report the results of NAS-Bench-201 on CIFAR-10~\cite{krizhevsky2009learning} dataset, TransNAS-Bench-101 Micro/Macro on Class Scene task, and NAS-Bench-Graph on Cora dataset in this part. The full results can be found in the Supplementary Material. The training portion in our experiments remains small, as many predictor-based NAS methods can already search top-performing architectures with very few training data~\cite{xu2021renas,white2021bananas,wei2022npenas}. Thus, studying performance under larger training sets is of limited practical significance. All results are averaged over 30 runs.
 
\noindent\textbf{Results on different search spaces.} We first focus on the top-ranking ability of each loss function. As shown in Figure~\ref{fig:res on search spaces}, weighted loss functions achieve the best Precision@0.5 and N@10 on all search spaces. For instance, MAPE takes the lead on NAS-Bench-201, and WARP ranks first on other search spaces. Conversely, MSE loss fails to recognize good architectures on the majority of search spaces. This failure can be attributed to their inability to rank architectures correctly~\cite{xu2021renas}. These results indicate that \textbf{weighted loss functions are good at identifying well-performing architectures} on both cell-based and skeleton-based search spaces.

For the overall ranking performance, ranking loss functions, such as HR, demonstrate a stable advantage on all search spaces. Moreover, we discover that training the predictor with a combination of two different loss functions simultaneously often results in worse results compared to using a single one. Specifically, the hybrid MSE+SR loss yields sub-optimal results, performing between regression and ranking loss functions across all three metrics.
\input{figs/mutation.tex}

\noindent\textbf{Trend on different training portions.} We further analyze how each metric evolves as the training portion increases. The third row of Figure~\ref{fig:res on training portions} illustrates a consistent improvement in Kendall Tau for most loss functions, indicating that more training data generally benefits the overall ranking ability of the predictor. However, the first and second rows of Figure~\ref{fig:res on training portions} reveal a counter-intuitive phenomenon: \textbf{more training data can degrade the top-ranking ability of certain loss functions}. This observation can be attributed to two key factors. First, regression and ranking loss functions treat all architectures equally rather than prioritizing well-performing ones, which inevitably leads to a decline in their ability to identify top architectures. Second, the quality of training data plays a crucial role, especially when only a very small subset of the search space is randomly sampled. For example, if a predictor is trained on just 0.1\% data, and this subset happens to contain several of the highest-performing architectures, it may develop a better ability to recognize promising architectures than a predictor trained on a larger but less representative dataset that primarily consists of poor architectures. 

Another fact is that \textbf{weighted loss functions generally perform better at distinguishing top-K architectures, but its effectiveness diminishes when the training dataset is extremely small}. We observe that WARP attains poor N@10 with the smallest training portion on NAS-Bench-201 and NAS-Bench-Graph. This is because limited training data causes weighted loss functions to overemphasize locally good architectures, assigning them disproportionately high weights. Hence, it fails to recognize good architectures in the entire search space. Ranking loss functions are more effective in such cases, as they leverage the overall ranking relationship to distinguish good architectures from a global perspective.

\noindent\textbf{Results on a mutation-based test set.} As suggested in prior work~\cite{white2021powerful}, it is also important to evaluate the performance of these loss functions to distinguish architectures which are local mutations from a few initial architectures. This is a more challenging scenario because architectures in the test set share similar structures. Specifically, we create a test set of 200 architectures by mutating top-10 seed architectures in the initial set which consists of 50 architectures randomly sampled from the search space. All the architectures in the test set differ seed architectures by one operation or edge. The results on NAS-Bench-201 are presented in Figure~\ref{fig:res on mutation}. We can observe a trend similar to Figure~\ref{fig:res on training portions} where the test set is the whole search space. Ranking loss functions, such as LR and HR, take the lead with limited training samples, while MAPE, a weighted loss function, shows superior performance when training samples increase. These results indicate that our findings still hold for neighborhood-based NAS methods such as evolutionary search and local search.
 
\input{figs/top_k.tex}

\noindent\textbf{Results on different levels of top-K/T.} We inspect the trend in Precision@T and N@K with an increasing K/T. As shown in Figure~\ref{fig:res on topk}, Precision@T and N@K of ranking loss functions are inferior to weighted ones, but the difference becomes smaller as K/T grows. For example, on NAS-Bench-201, WARP outperforms HR in Precision@0.5 by 2.85\% but achieves a 1.21\% lower Precision@5 than HR. This is because the overall ranking ability matters more when K/T becomes larger, which aligns with the optimization object of ranking loss functions.   

\input{figs/weights.tex}

\input{tabs/ap_and_pinat.tex}

\input{figs/search_res.tex}
\noindent\textbf{Impacts of weighting types in weighted loss functions.} To investigate how weighting types influence the effectiveness of weighted loss functions, we use three types of weights (`GT': $\hat{y}$; `EXP-GT': $\exp(\hat{y})$; `Ranking': $1-(r-1)/n$, $r$ denotes the actual ranking) for WARP and EW on NAS-Bench-201. Figure~\ref{fig:weights} shows that `EXP-GT' achieves the best top-ranking precision and `GT' yields close results. Both of them outperform `Ranking' by a large margin. These results indicate that \textbf{involving GT performances of architectures in the weight is generally a better choice for weighted loss functions than involving their rankings.} This is because GT could reflect the global importance of the architecture, while the ranking only focuses on its importance in the limited training data locally. The former helps weighted loss functions generalize better on the entire search space. 

\noindent\textbf{Results on different types of predictors.} We examine representative loss functions on another two performance predictors: the Multi-Layer Perceptron (MLP)-based Accuracy Predictor (AP)~\cite{cai2019once} and the Transformer-based PINAT~\cite{lu2023pinat}. The results on NAS-bench-201 are shown in Table~\ref{table:ap and pinat}. We can see that AP performs best with ListMLE in all metrics, while WARP helps PINAT achieve the lowest N@10 and the highest Precision@0.5. The inferior performance of WARP for AP can be credited to its MLP backbone, which can easily overfit the limited training data and fall into the local optimal with the weighted loss function. Conversely, the Transformer backbone helps PINAT generalize well on the whole search space. In terms of identifying well-performing architectures, these results reflect that \textbf{predictors with a simple backbone such as MLP work better with ranking loss functions, while predictors with a generalizable backbone such as GCN and Transformer achieve better results with weighted loss functions.}

\input{tabs/search_201.tex}

\subsection{Predictor-based NAS Evaluation}
Now, we investigate the impact of different types of loss functions in predictor-based NAS. In this set of experiments, we employ the popular predictor-guided evolutionary search~\cite{jing2022graph}, which trains the predictor many times during the iterative sampling. Since DARTS~\cite{liu2018darts} does not provide the architecture accuracy, we train architectures for 50 epochs and evaluate them on the validation dataset as the training samples for the predictor. We also propose a new predictor-based NAS algorithm and compare its effectiveness with prior works. The results are averaged over multiple runs (DARTS for 5 and others for 20). 

\input{tabs/search_101.tex}

\paragraph{Predictor-based NAS with a piecewise loss function.} One significant finding from Figure~\ref{fig:res on training portions} is that different types of loss functions are specialized under specific training portions. Based on this, we improve the predictor-based NAS by changing its fixed loss function to a flexible PW loss. We call this new method PWLNAS. Specifically, we use regression/ranking loss functions in early iterations to warm up the predictor and then shift to the weighted one to recognize good architectures. The choice of loss functions and the number of warm-up iterations depends on the specific task. We give detailed settings for PWLNAS as well as the pseudo-code of PWLNAS in the Supplementary Material.

\noindent\textbf{Searching on NAS-Bench-201.}
 We use a PW loss composed of HR and MAPE. Comparisons between different loss functions are reported in Figure~\ref{fig:search res 201}. We see that PW loss achieves the lowest test error rate than any single one on three datasets. HR performs better than MAPE in the region of low query budget, which is consistent with our finding in Section~\ref{sec: predictor evaluation}. Besides, as shown in Table~\ref{table:search 201}, PWLNAS outperforms competitive predictor-based methods. In particular, PWLNAS can find the best architecture on the CIFAR-10 dataset. 

\noindent\textbf{Searching on NAS-Bench-101.}
 We use a PW loss composed of ListMLE and WARP. Table~\ref{table:search 101} summarizes a major advantage of PW loss over single ones. Additionally, PWLNAS also surpasses previous SOTA methods like NPENAS~\cite{wei2022npenas} and FlowerFormer~\cite{hwang2024flowerformer}. Note that such improvement is non-trivial compared with those improvements claimed in prior works. 

\input{tabs/search_micro.tex}
\input{tabs/darts_cf10.tex}

\noindent\textbf{Searching on TransNAS-Bench-101 Micro.}
We use a PW loss composed of MSE and EW for the Jigsaw task as well as a PW loss composed of HR and WARP for other tasks. According to Table~\ref{table: micro search}, PWLNAS takes the lead on all tasks with PW loss when compared with competitive methods. Besides, several loss functions only perform well on specific tasks. For example, MSE ranks second on Jigsaw  but performs worst on Class Scene. This indicates the necessity of effectively combining different types of loss functions to improve predictor-based NAS.

\noindent\textbf{Searching on DARTS.}
We use a PW loss composed of HR and MAPE. Table~\ref{table:darts cf10} demonstrates the results on DARTS with 100 queried architectures. We find that PWLNAS achieves the lowest test error rate of 2.47$\%$, beating prior SOTA methods like DCLP~\cite{zheng2024dclp}.

%% file: figs/training_portion.tex
\begin{figure*}[t]
     \begin{minipage}[b]{0.19\textwidth}
        \centering
        \includegraphics[width=\textwidth]{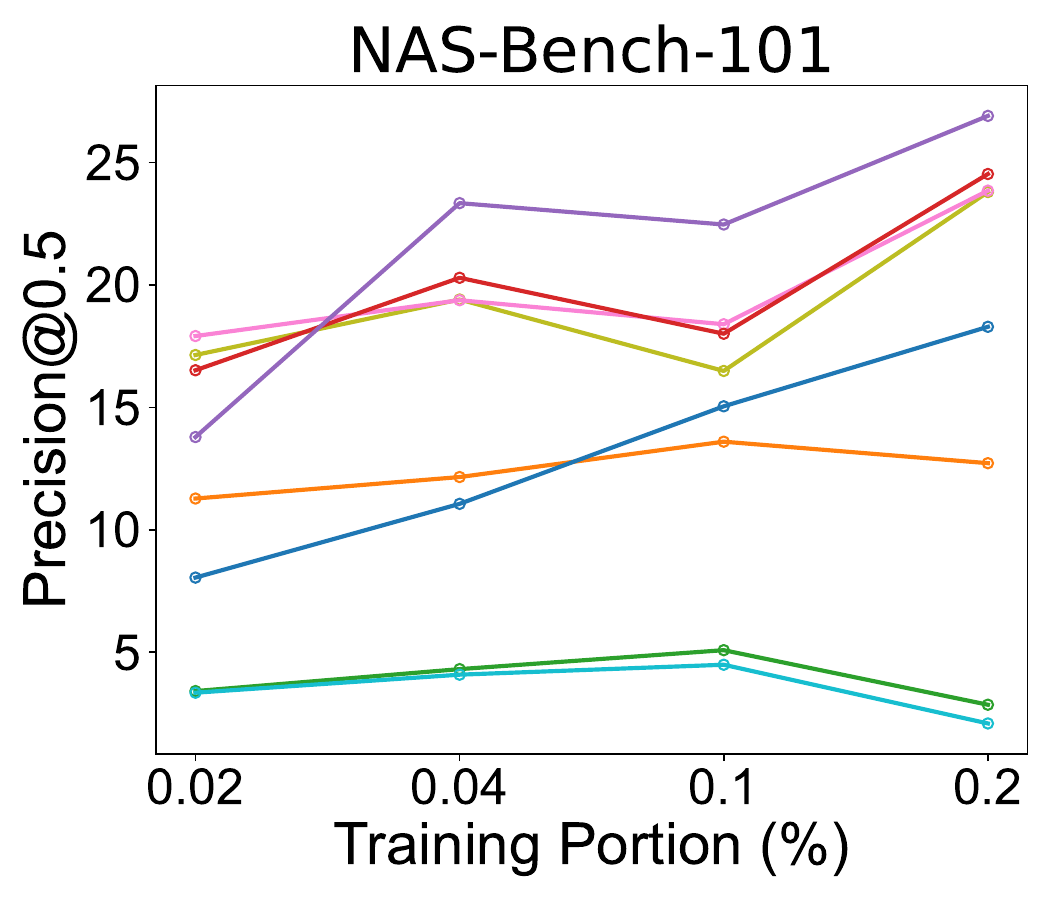}
    \end{minipage}\hfill   
     \begin{minipage}[b]{0.19\textwidth}
        \centering
        \includegraphics[width=\textwidth]{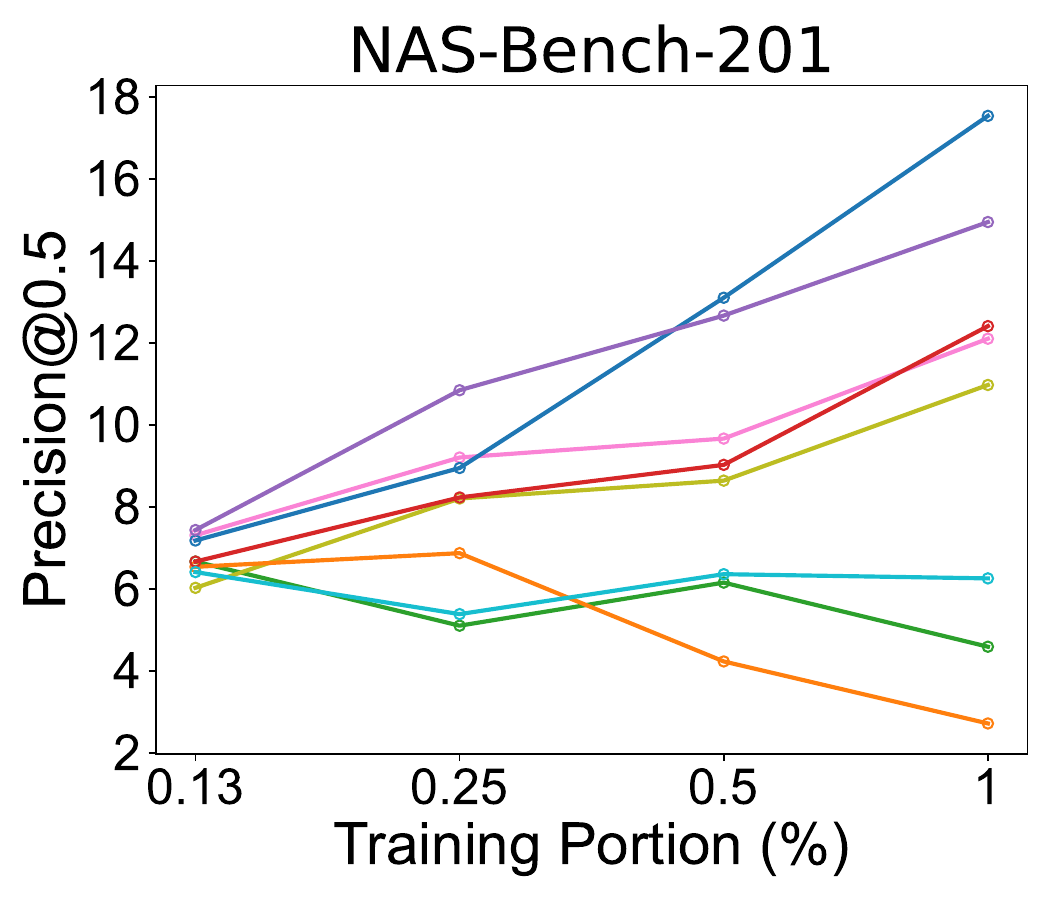}
    \end{minipage}\hfill   
    \begin{minipage}[b]{0.19\textwidth}
        \centering
        \includegraphics[width=\textwidth]{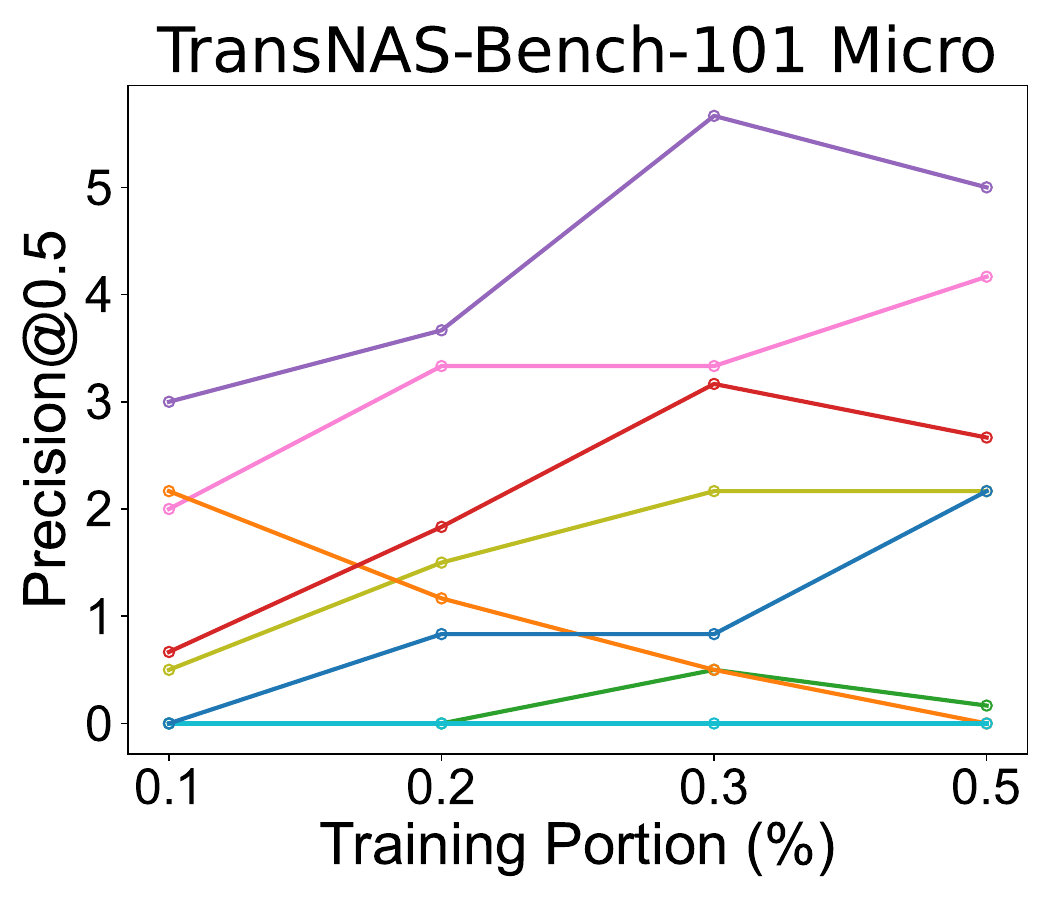}
    \end{minipage}\hfill
      \begin{minipage}[b]{0.19\textwidth}
        \centering
        \includegraphics[width=\textwidth]{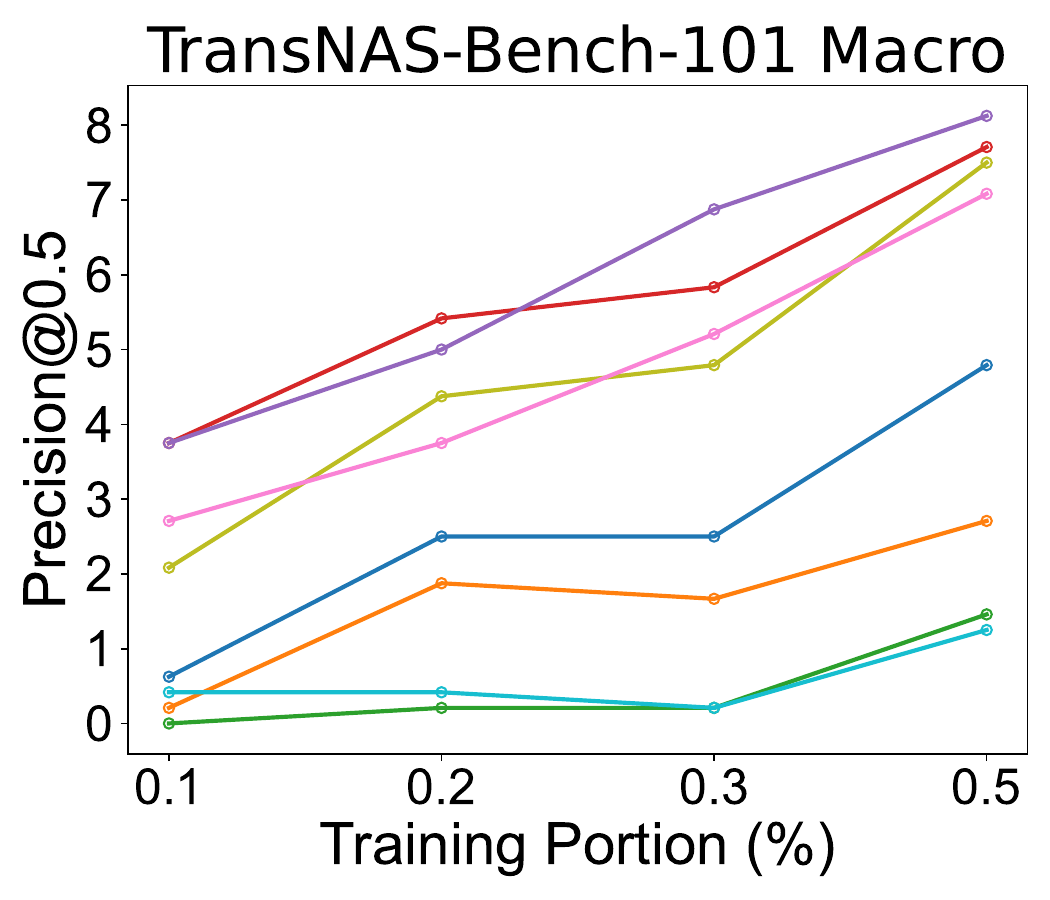}
    \end{minipage}\hfill   
      \begin{minipage}[b]{0.19\textwidth}
        \centering
        \includegraphics[width=\textwidth]{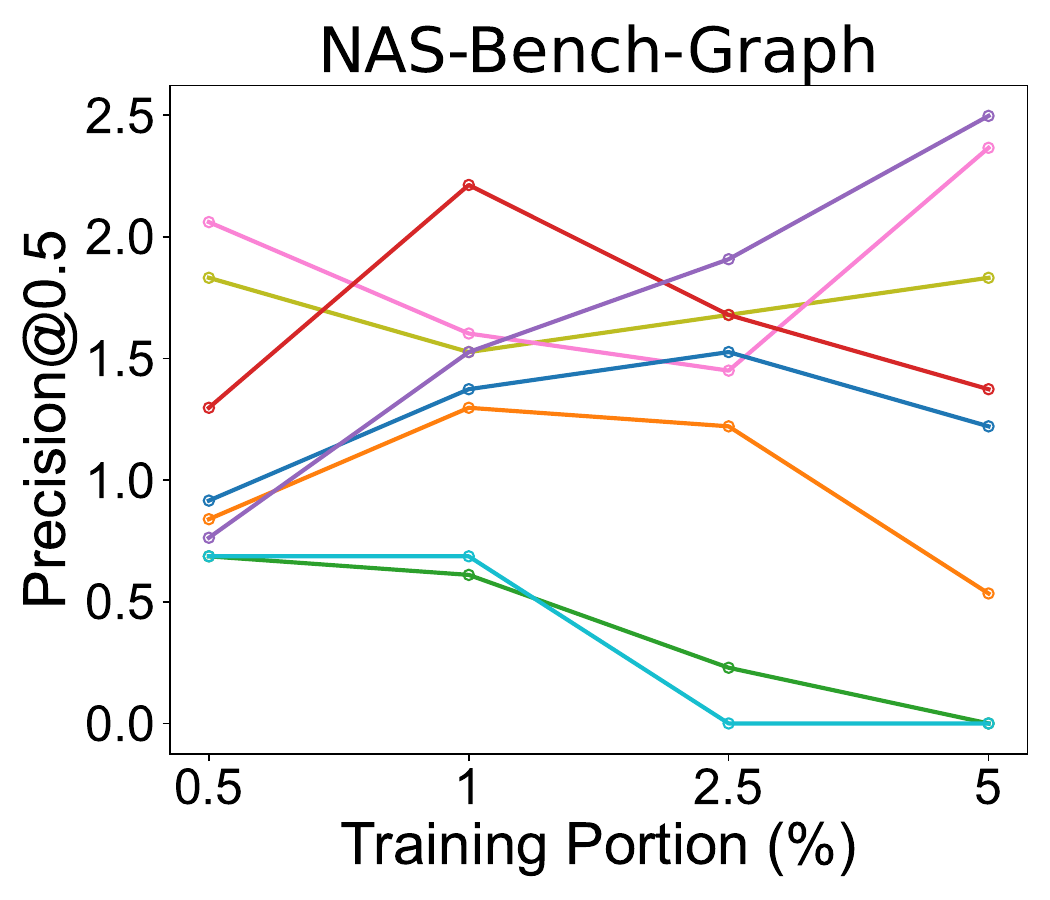}
    \end{minipage}   
    \begin{minipage}[b]{0.19\textwidth}
        \centering
        \includegraphics[width=\textwidth]{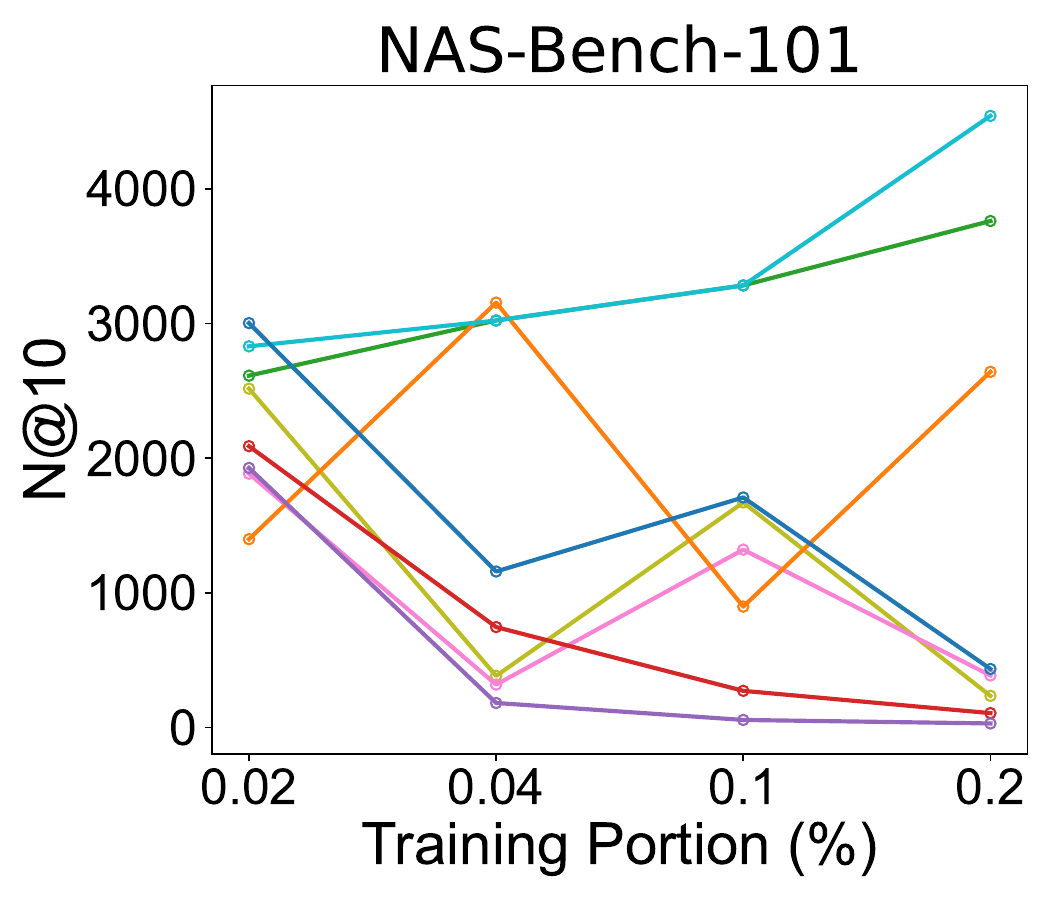}
    \end{minipage}\hfill   
     \begin{minipage}[b]{0.19\textwidth}
        \centering
        \includegraphics[width=\textwidth]{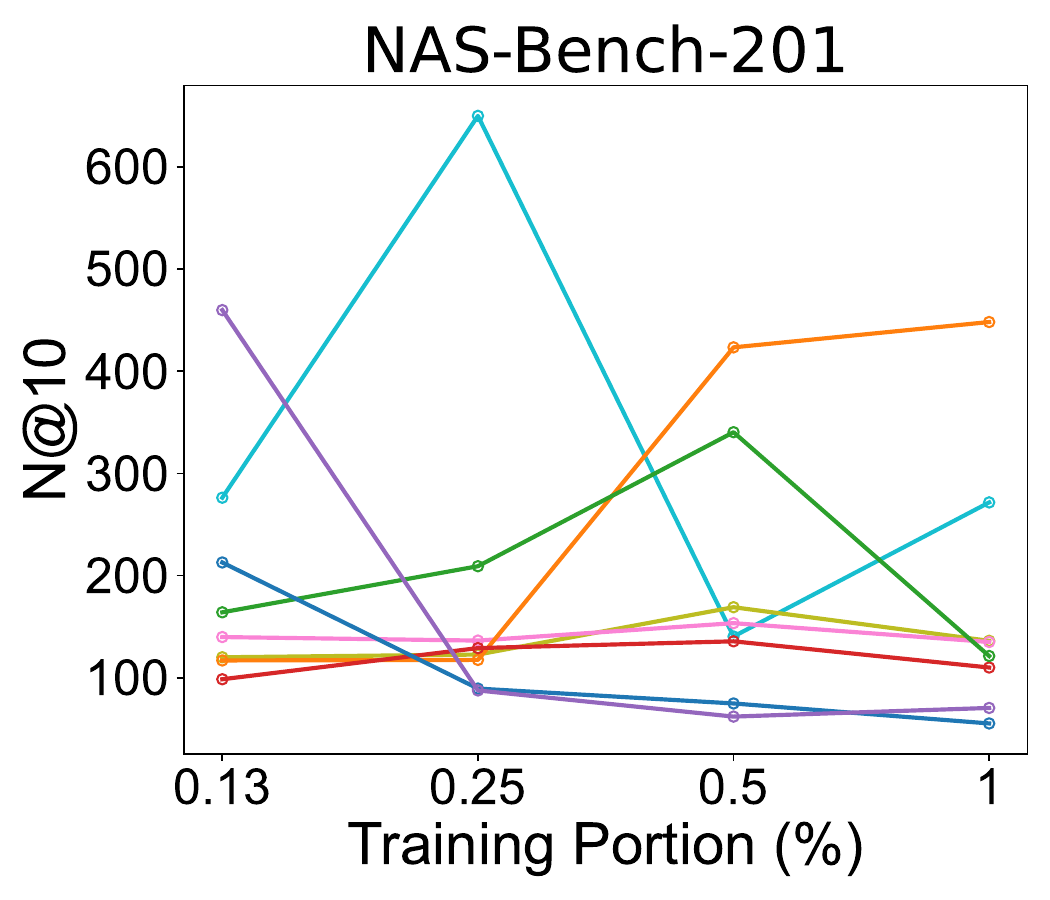}
    \end{minipage}\hfill   
    \begin{minipage}[b]{0.19\textwidth}
        \centering
        \includegraphics[width=\textwidth]{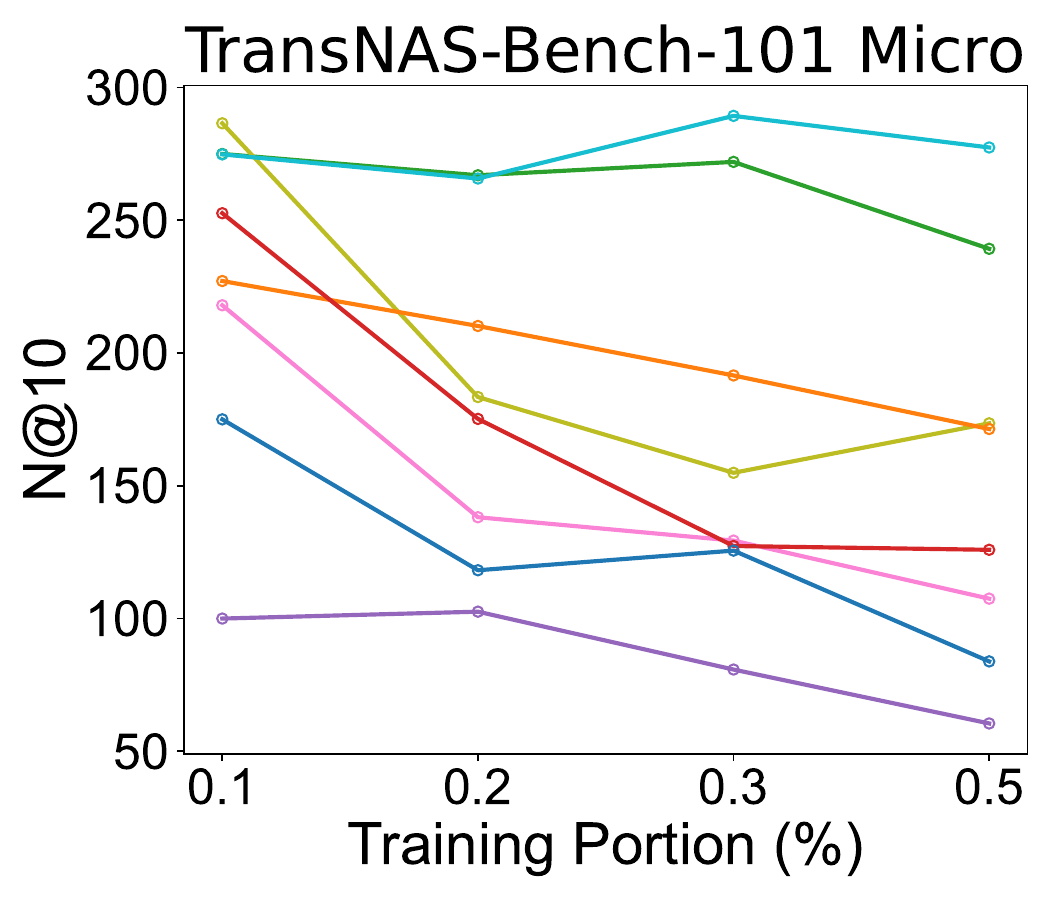}
    \end{minipage}\hfill
      \begin{minipage}[b]{0.19\textwidth}
        \centering
        \includegraphics[width=\textwidth]{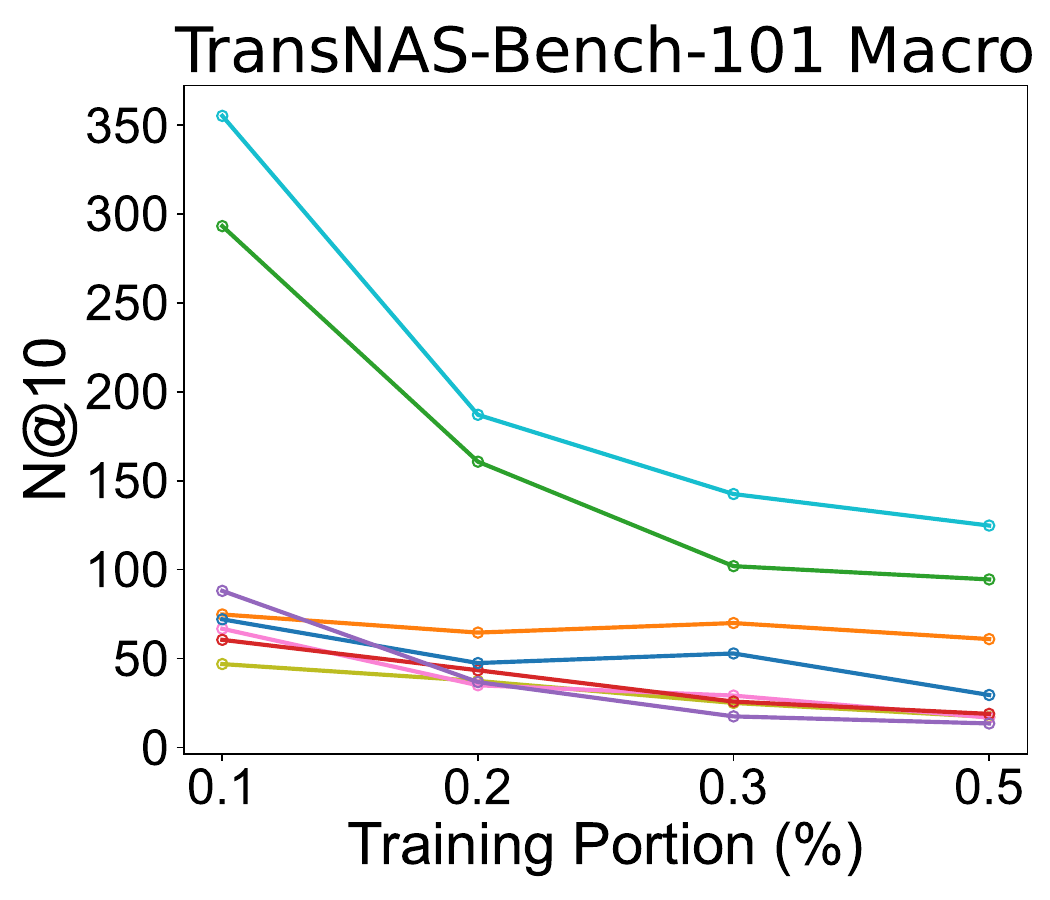}
    \end{minipage}\hfill
      \begin{minipage}[b]{0.19\textwidth}
        \centering
        \includegraphics[width=\textwidth]{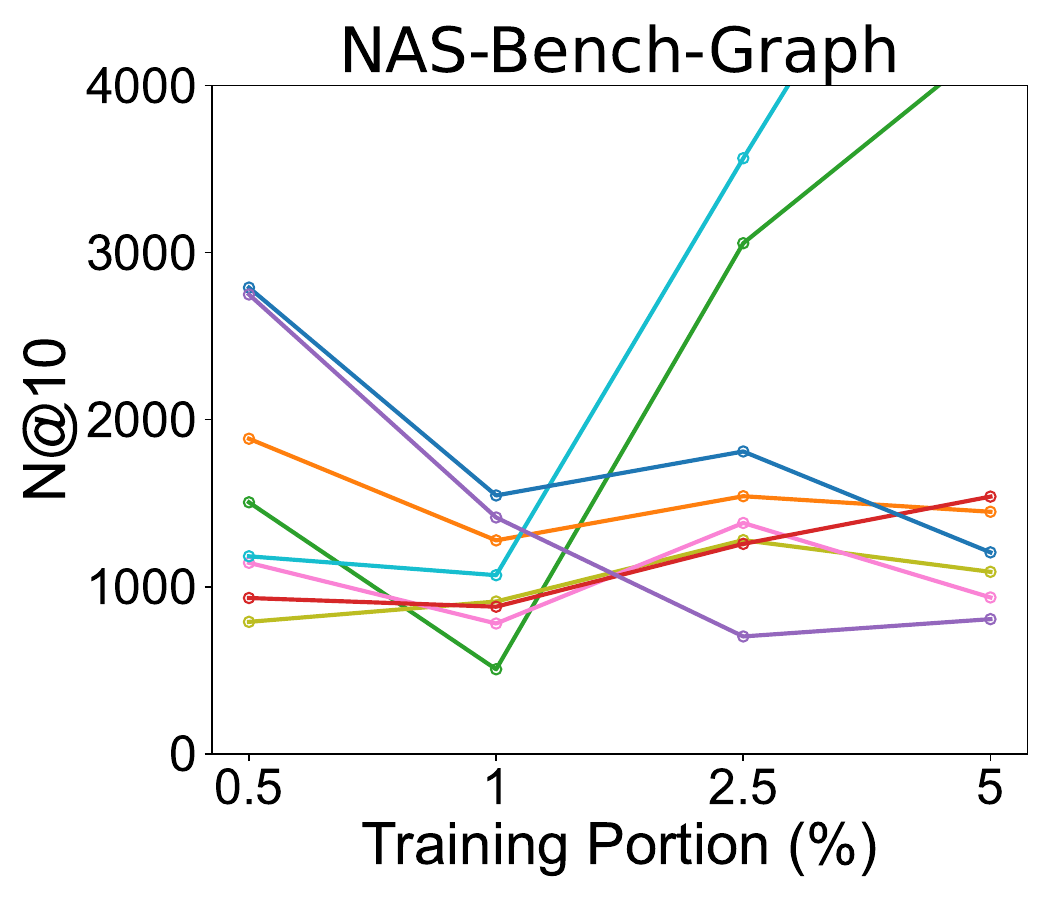}
    \end{minipage}   
    \begin{minipage}[b]{0.19\textwidth}
        \centering
        \includegraphics[width=\textwidth]{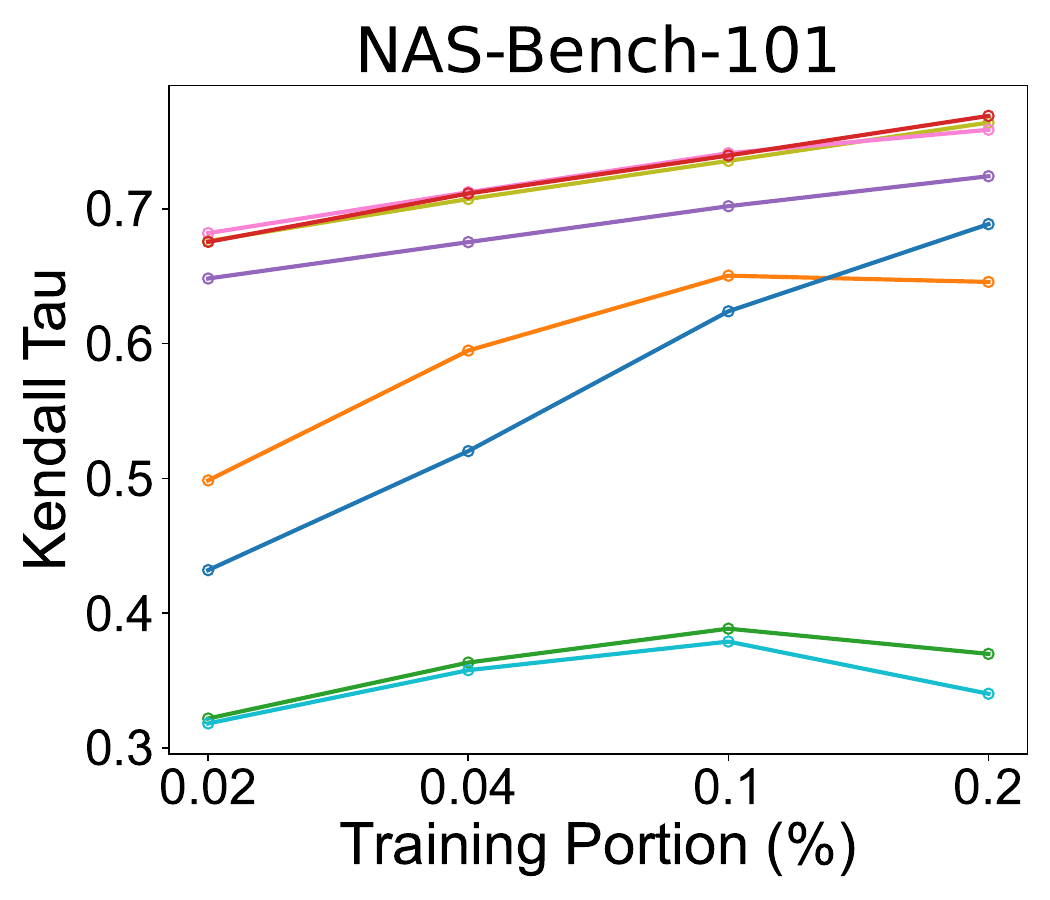}
    \end{minipage}\hfill   
     \begin{minipage}[b]{0.19\textwidth}
        \centering
        \includegraphics[width=\textwidth]{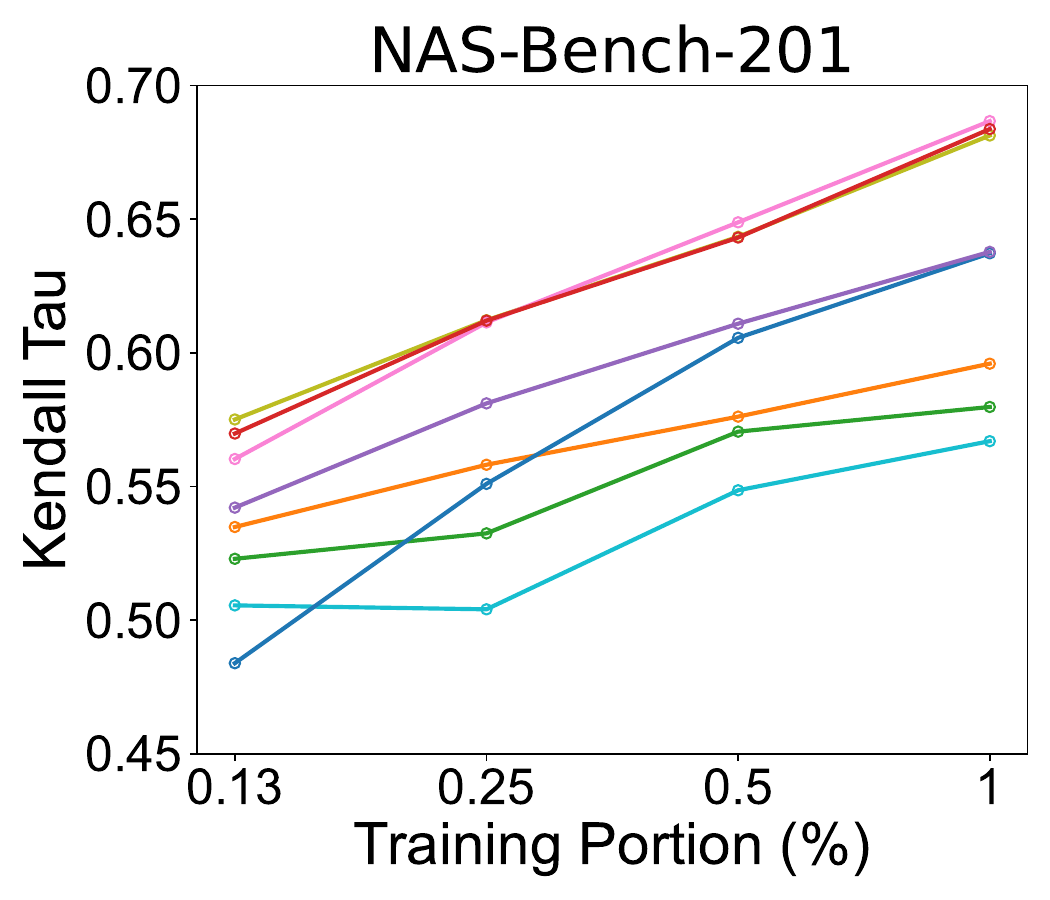}
    \end{minipage}\hfill   
    \begin{minipage}[b]{0.19\textwidth}
        \centering
        \includegraphics[width=\textwidth]{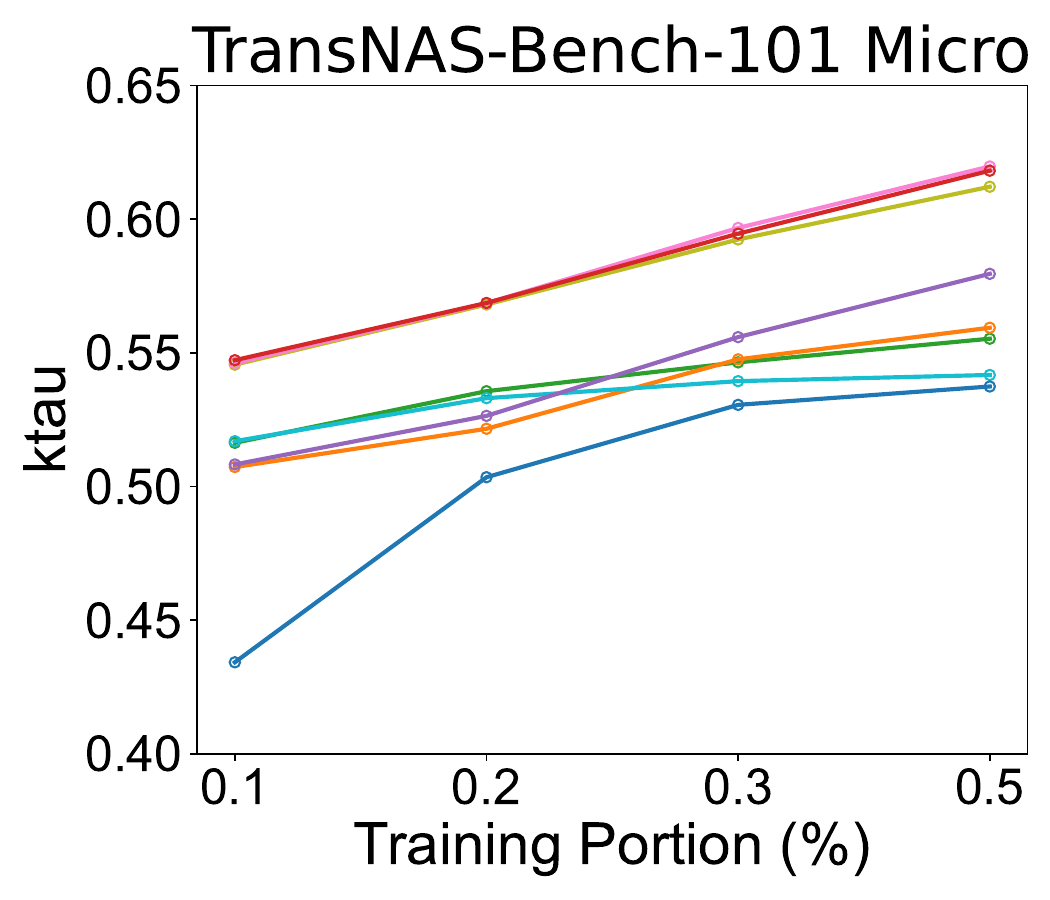}
    \end{minipage}\hfill
      \begin{minipage}[b]{0.19\textwidth}
        \centering
        \includegraphics[width=\textwidth]{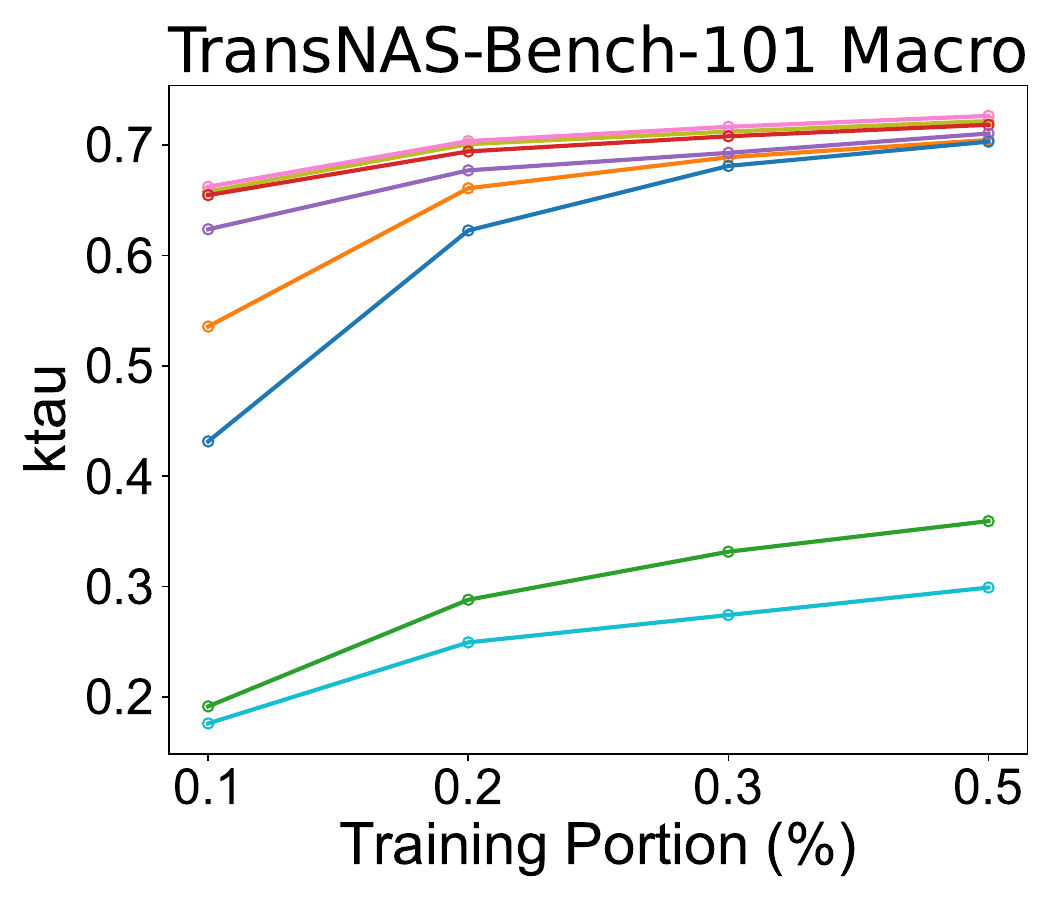}
    \end{minipage}\hfill   
      \begin{minipage}[b]{0.19\textwidth}
        \centering
        \includegraphics[width=\textwidth]{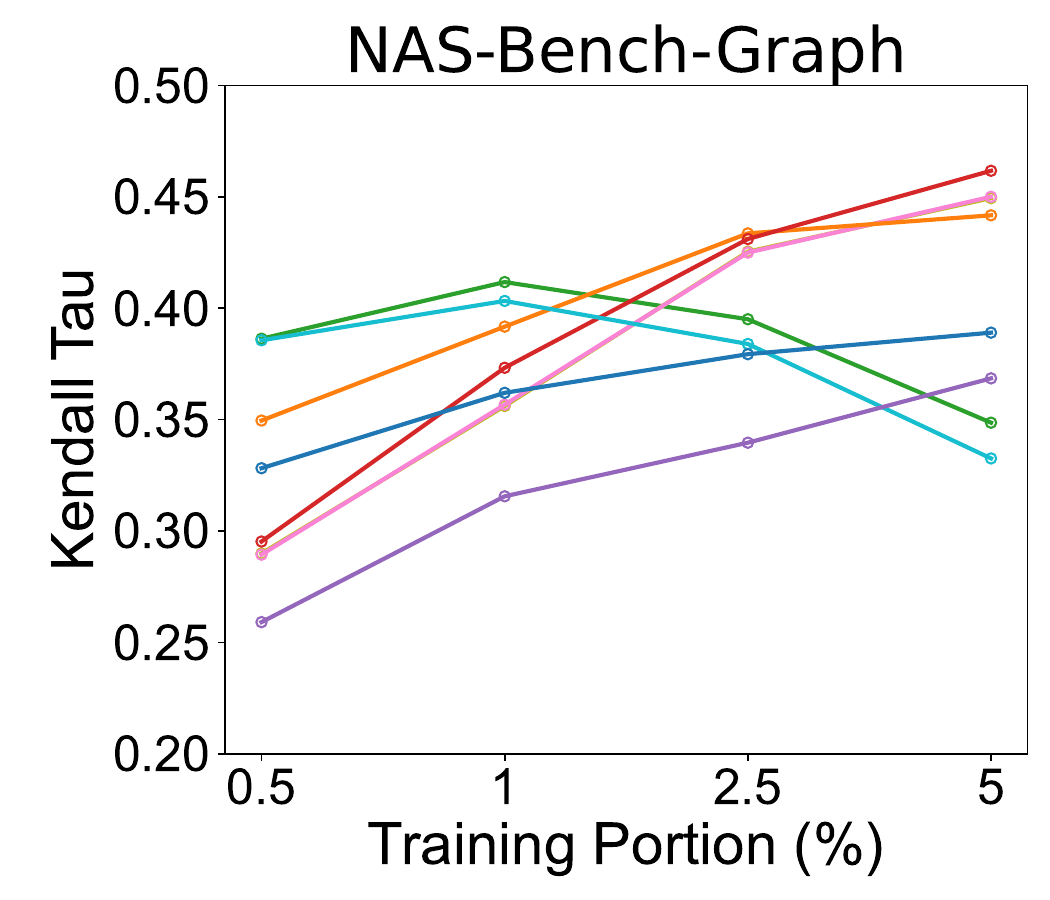}
    \end{minipage}
    \begin{minipage}[b]{0.95\textwidth}
        \centering
         \hspace{20cm}\includegraphics[width=\textwidth]{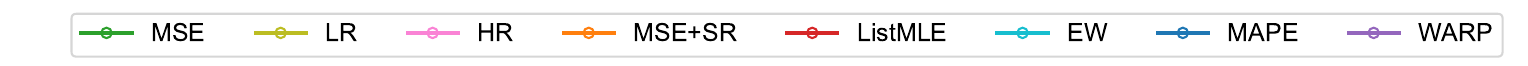}
    \end{minipage}
     \caption{Precision@0.5, N@10, and Kendall's Tau (\textit{Top to Bottom}) of different loss functions. Note that a higher metric indicates better performance except for N@10.}
    \label{fig:res on training portions}
\end{figure*}

%% file: figs/mutation.tex
\begin{figure}[t]
   \begin{minipage}[b]{0.23\textwidth}
        \centering
        \includegraphics[width=\textwidth]{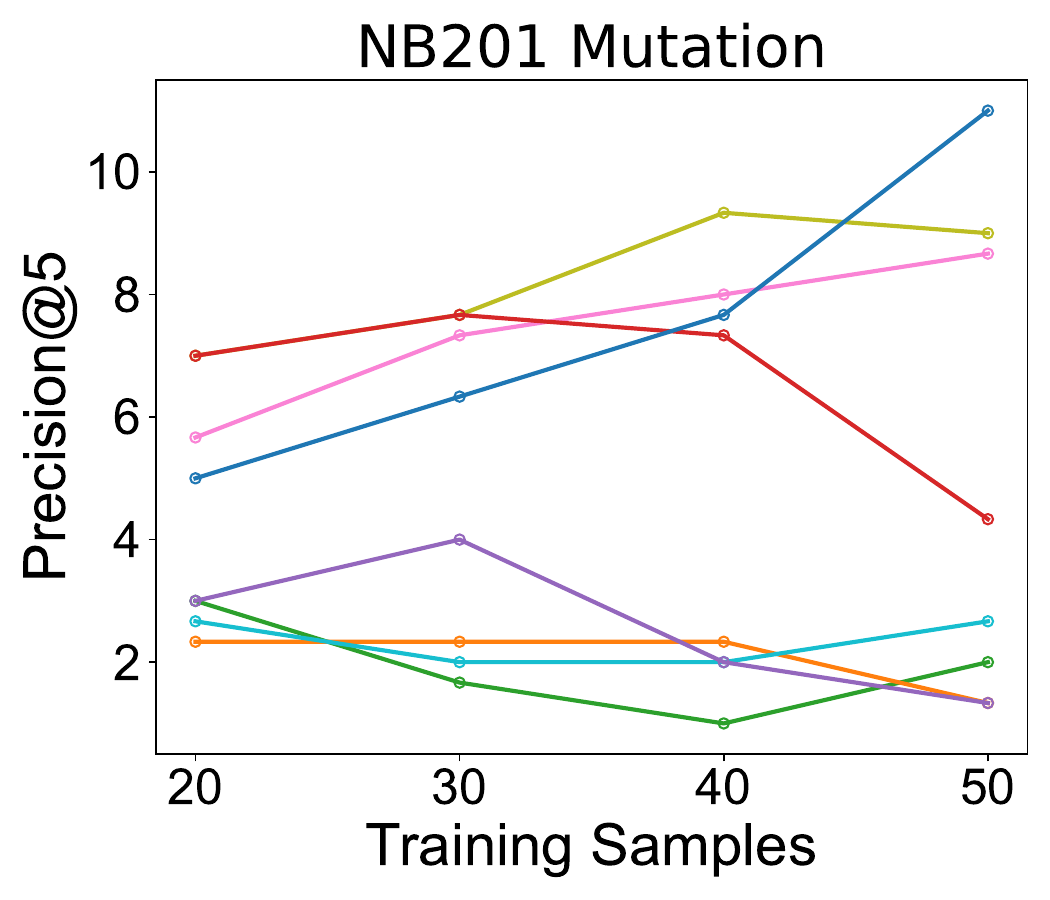}
    \end{minipage}\hfill
    \begin{minipage}[b]{0.23\textwidth}
        \centering
        \includegraphics[width=\textwidth]{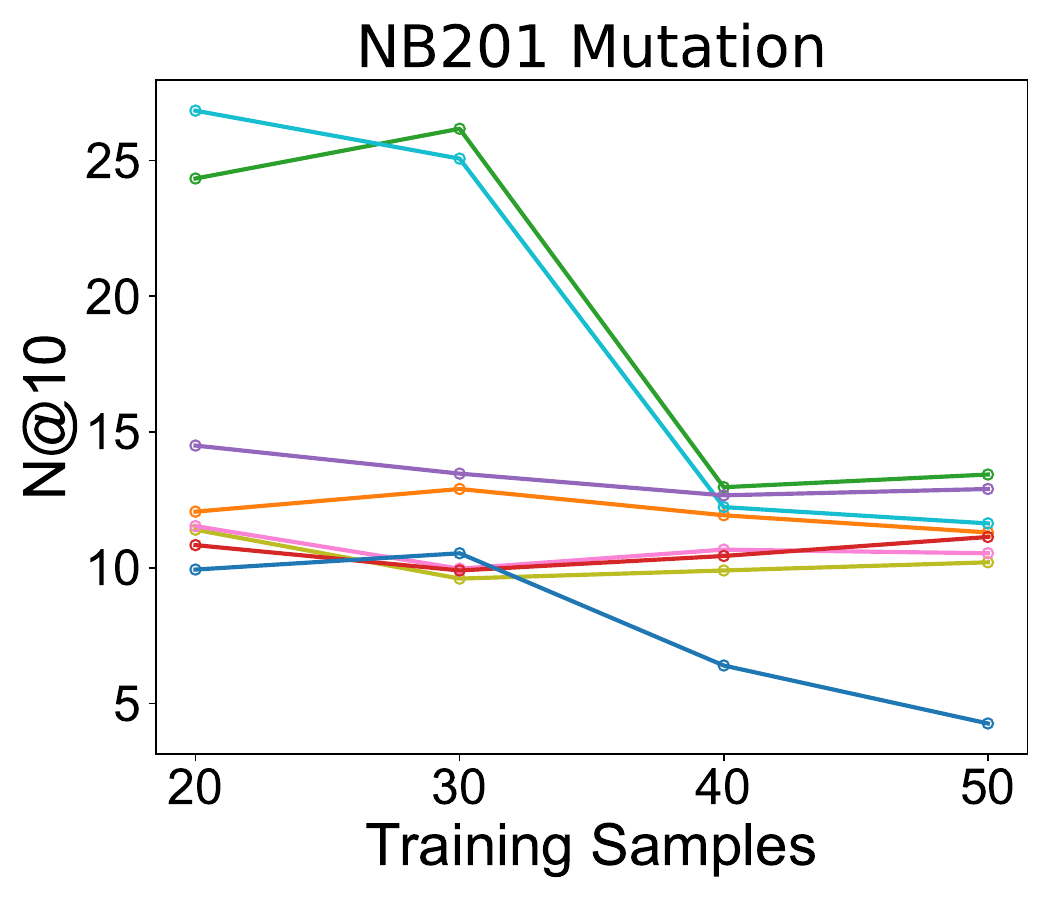}
    \end{minipage}\hfill   
    \begin{minipage}[b]{0.45\textwidth}
        \centering
         \hspace{20cm}\includegraphics[width=\textwidth]{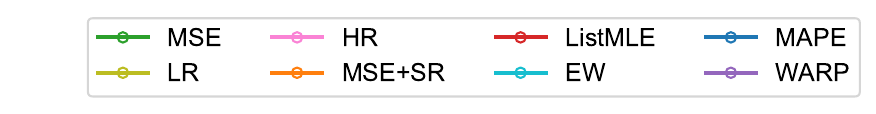}
    \end{minipage}
     \caption{Precision@5 and N@10 of different loss functions for a mutation-based test set on NAS-Bench-201. Note that a higher Precision@5 and a lower N@10 are preferred.}
    \label{fig:res on mutation}
\end{figure}

%% file: figs/top_k.tex
\begin{figure}[t]
   \begin{minipage}[b]{0.23\textwidth}
        \centering
        \includegraphics[width=\textwidth]{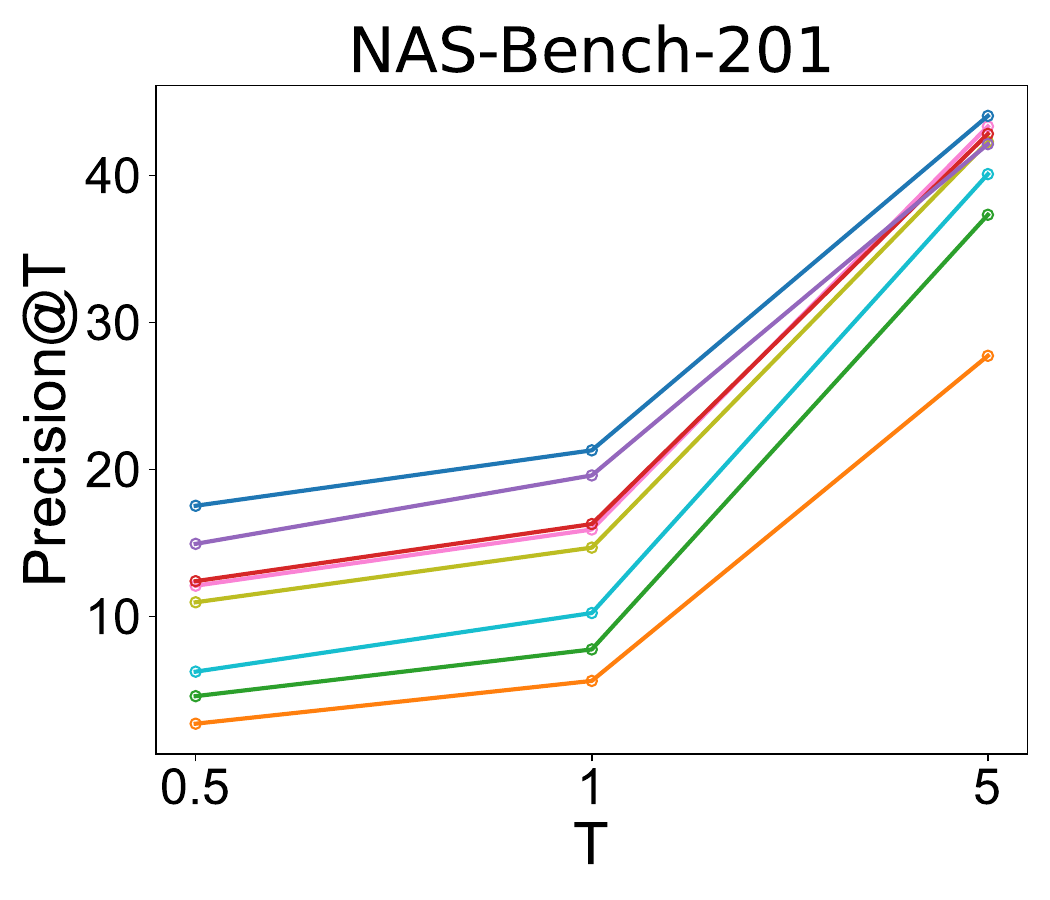}
    \end{minipage}\hfill
      \begin{minipage}[b]{0.23\textwidth}
        \centering
        \includegraphics[width=\textwidth]{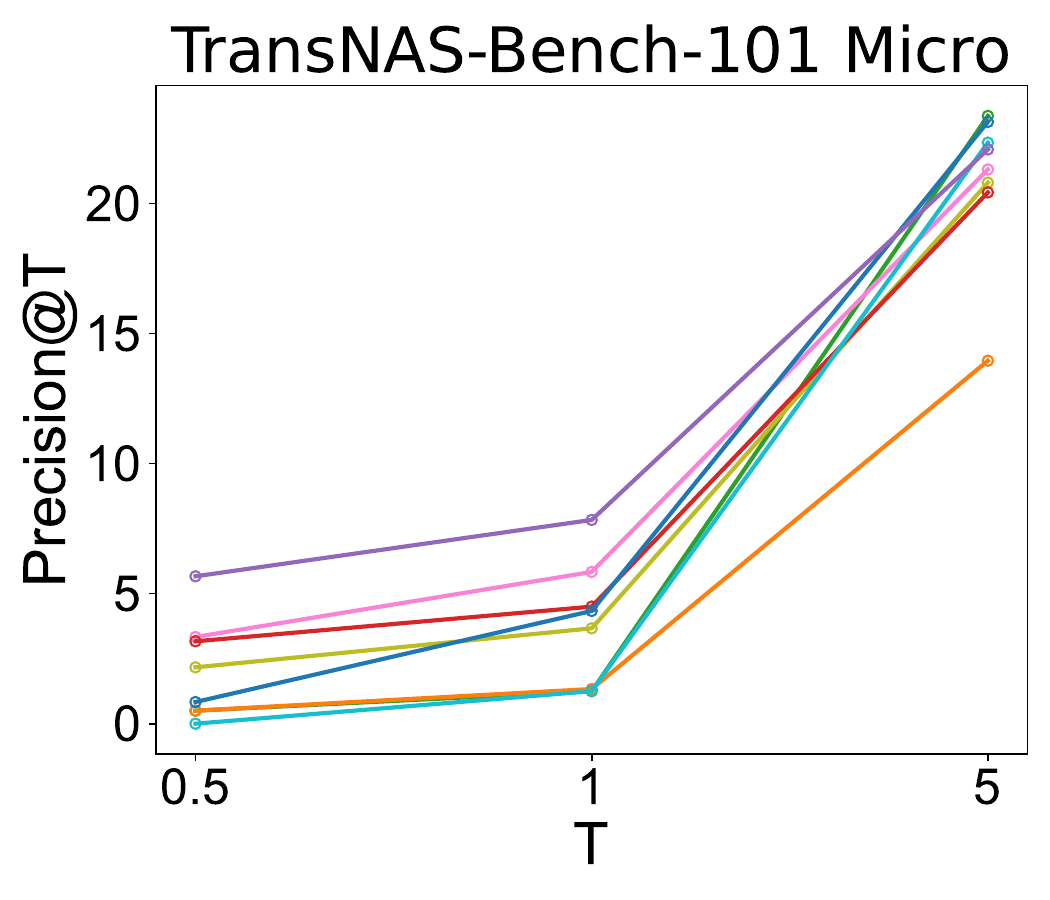}
    \end{minipage}\hfill  
    \vspace{-0.5em}
    \begin{minipage}[b]{0.23\textwidth}
        \centering
        \includegraphics[width=\textwidth]{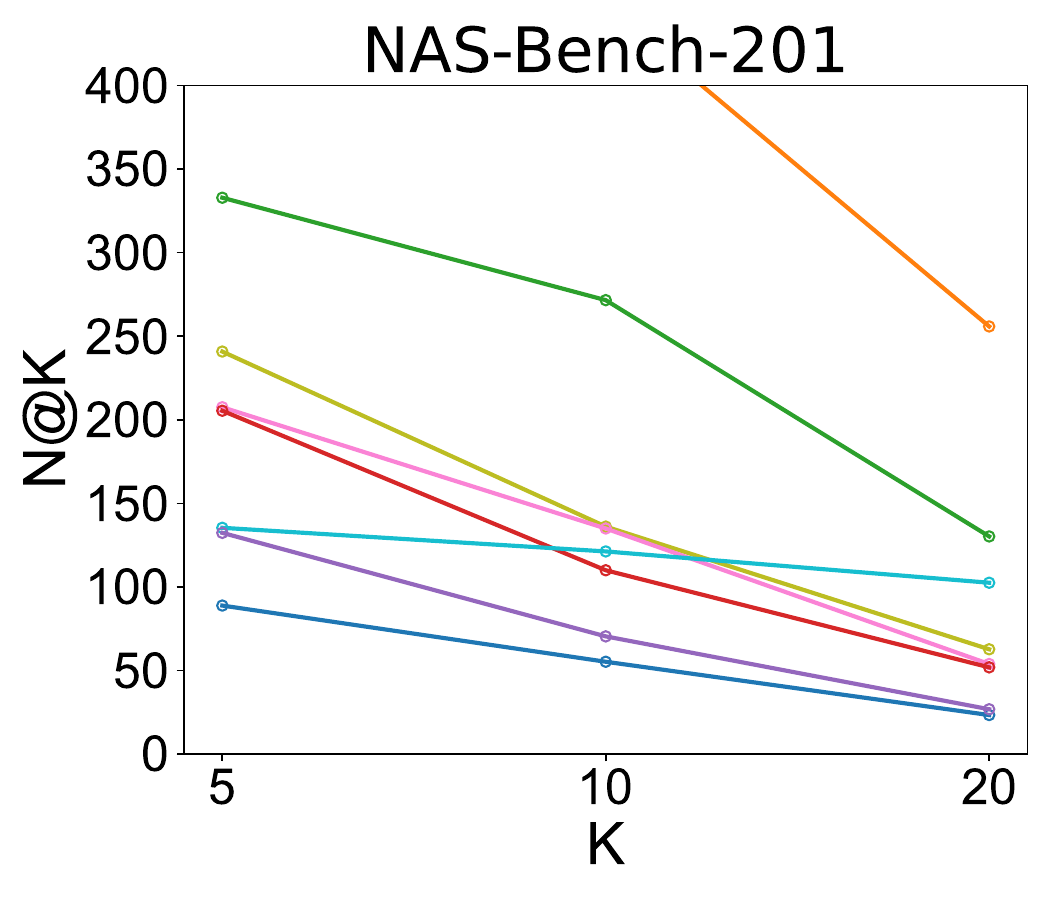}
    \end{minipage}\hfill   
     \begin{minipage}[b]{0.23\textwidth}
        \centering
        \includegraphics[width=\textwidth]{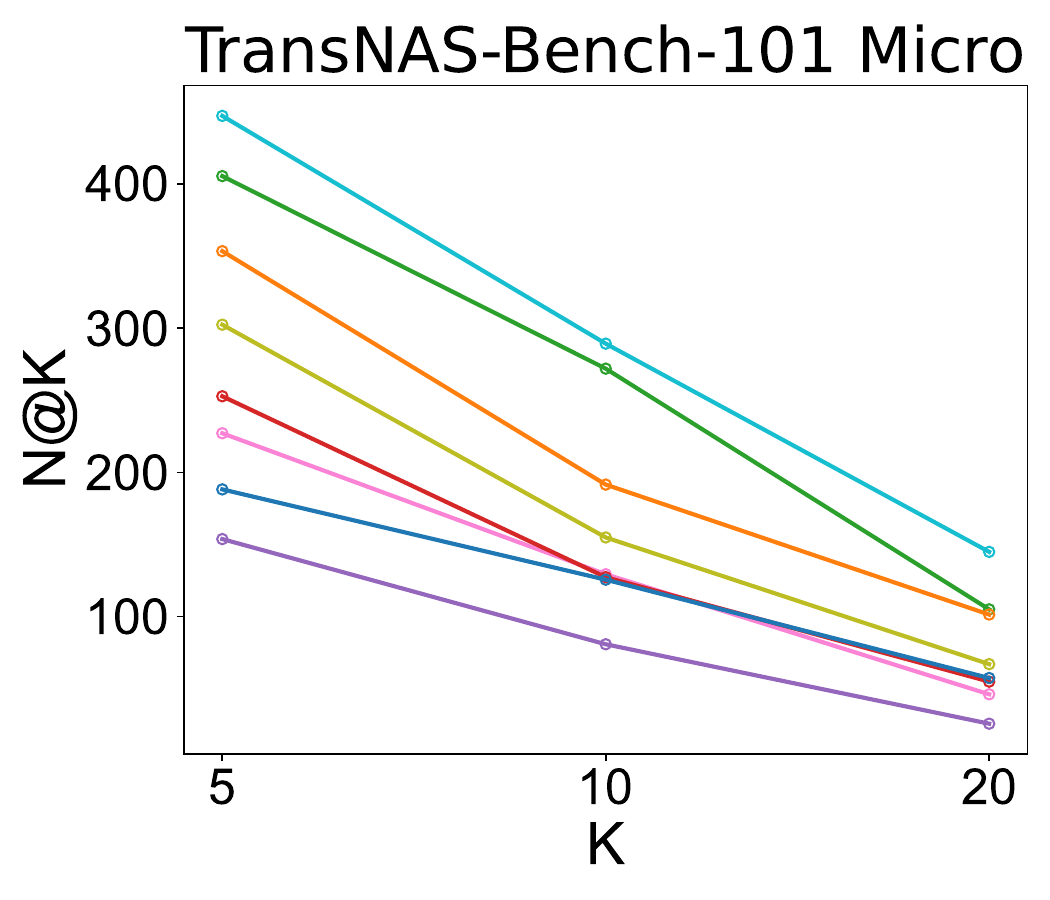}
    \end{minipage}\hfill   
    \vspace{-0.5em}
    \vspace{-0.5em}
    \begin{minipage}[b]{0.45\textwidth}
        \centering
         \hspace{20cm}\includegraphics[width=\textwidth]{images/legend_plot_4.pdf}
    \end{minipage}
     \caption{Precision@T and N@K of different loss functions on NAS-Bench-201 with 1\% training data and TransNAS-Bench-101 Micro with 0.3\% training data. Note that a higher Precision@T and a lower N@K is preferred.}
    \label{fig:res on topk}
\end{figure}

%% file: figs/weights.tex
\begin{figure}[t]
     \begin{minipage}[b]{0.23\textwidth}
        \centering
        \includegraphics[width=\textwidth]{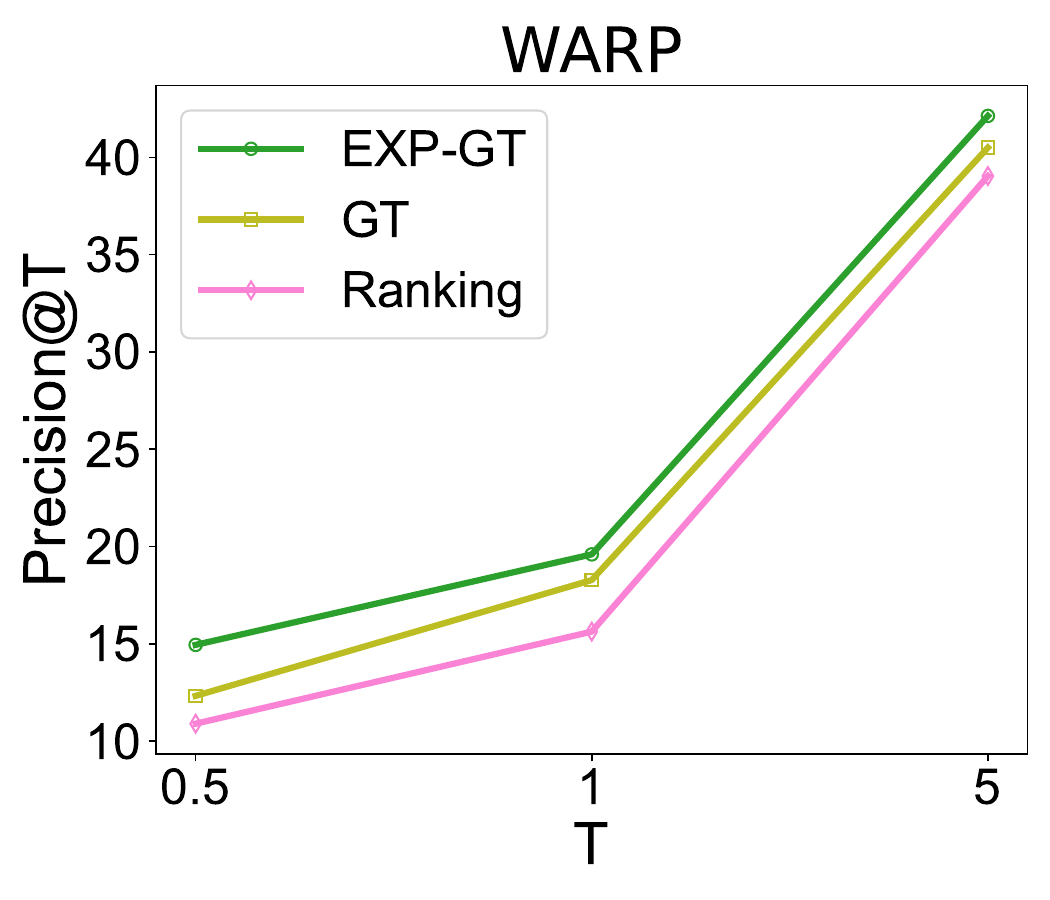}
    \end{minipage}\hfill   
     \begin{minipage}[b]{0.23\textwidth}
        \centering
        \includegraphics[width=\textwidth]{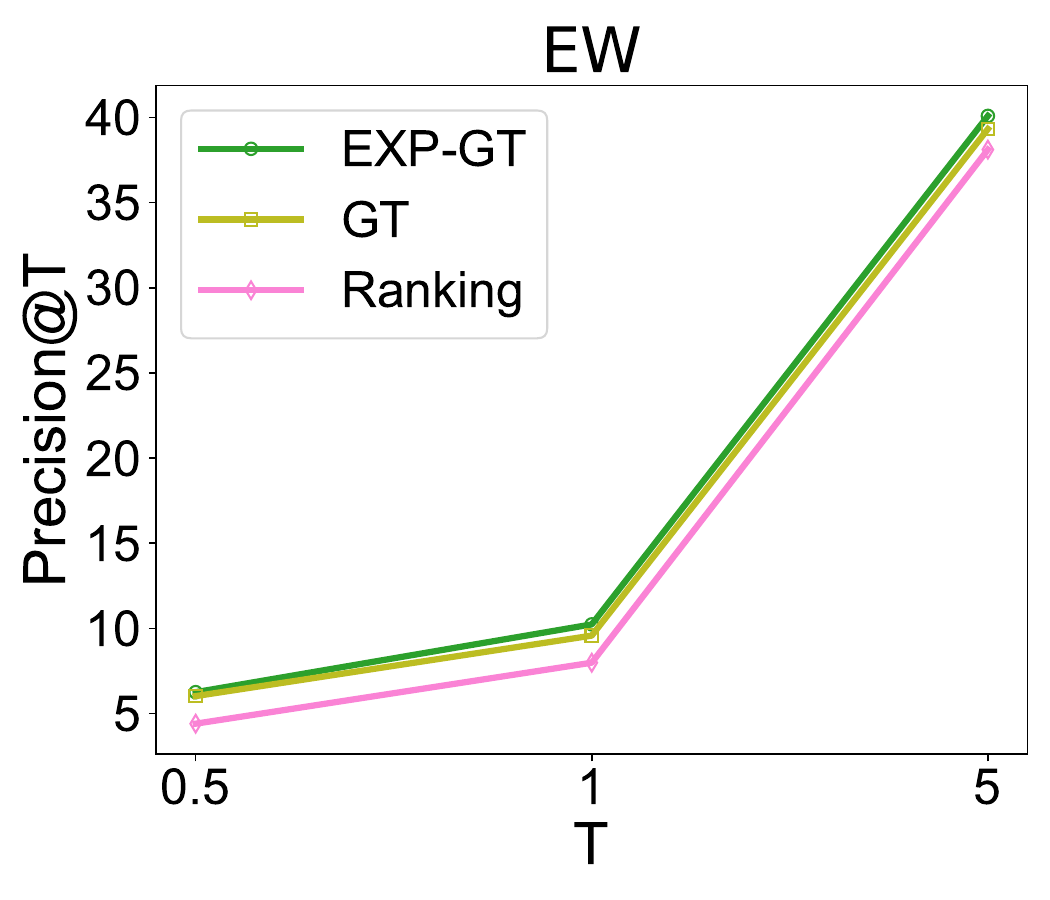}
    \end{minipage}\hfill 
    \vspace{-0.5em}
     \begin{minipage}[b]{0.23\textwidth}
        \centering
        \includegraphics[width=\textwidth]{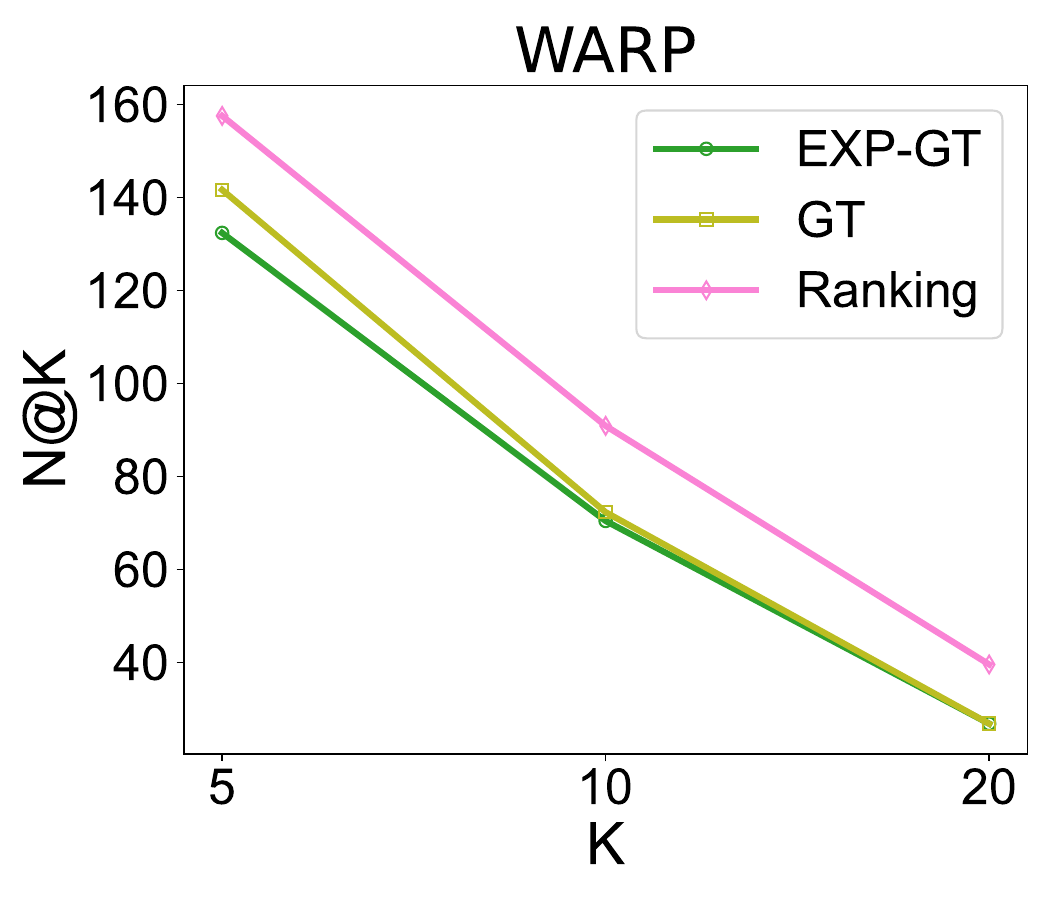}
    \end{minipage}\hfill   
     \begin{minipage}[b]{0.23\textwidth}
        \centering
        \includegraphics[width=\textwidth]{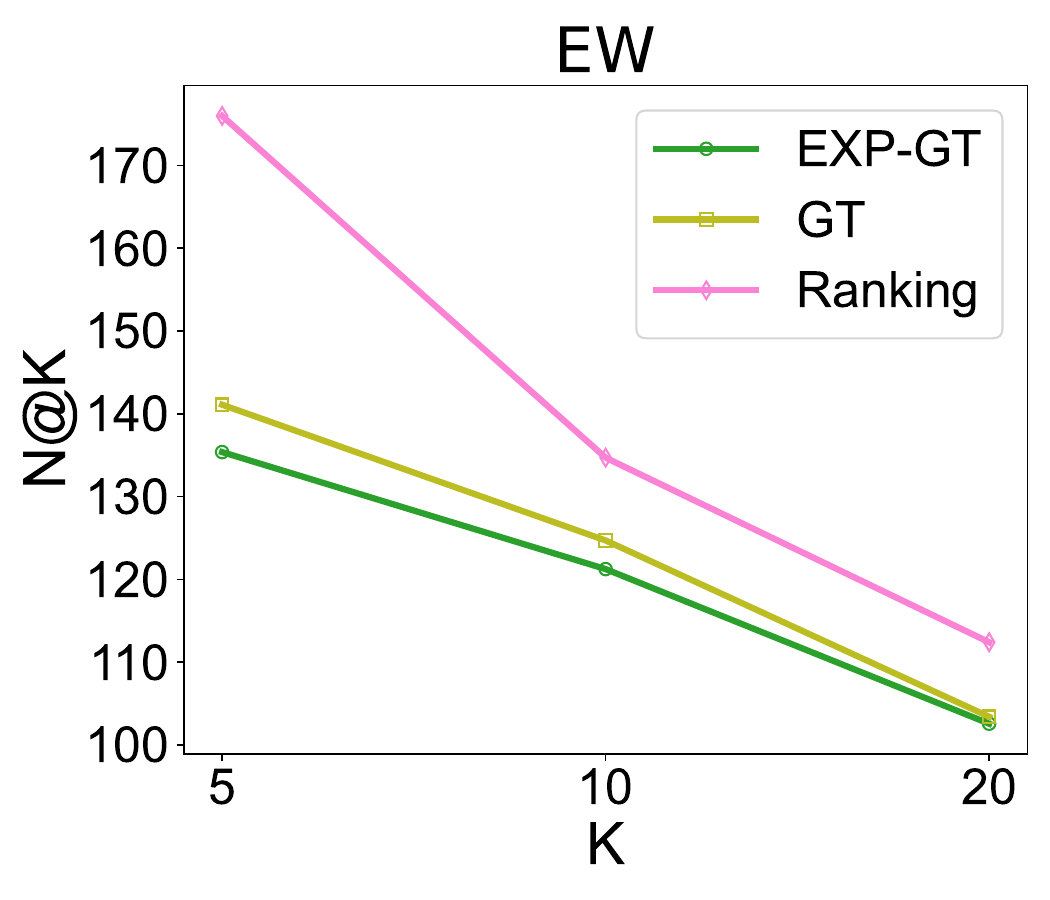}
    \end{minipage}\hfill 
     \caption{Precision@T and N@K of different weighting types for WARP and EW on NAS-Bench-201 with 1\% training data. Note that a higher Precision@T and a lower N@K are preferred.}
    \label{fig:weights}
\end{figure}

%% file: tabs/ap_and_pinat.tex
\begin{table}[t]
 \centering
   \renewcommand{\arraystretch}{1.13}
\caption{Comparisons of other predictors with 1\% training data on NAS-Bench-201. Precision@0.5 is short for Ptop@0.5. \textbf{Bold} indicates the best.}
\vspace{-0.5em}
 \resizebox{0.98\columnwidth}{!}{
        \begin{tabular}{c|c|c|c|c|c}
        \thickhline
        Predictor & Backbone  &Loss  & N@10$^\downarrow$   & Ptop@0.5$^\uparrow$  & $\tau^{\uparrow}$\\
         \hline
         \multirow{4}{*}{AP~\cite{cai2019once}} &\multirow{4}{*}{MLP}
        & MSE  & 250.94 & 4.41  & 0.43  \\
        & & HR  & 23.58 & 22.15    & 0.65  \\
        & & ListMLE  & \textbf{22.74} & \textbf{24.15}  & \textbf{0.66}                        \\
        & & WARP  & 113.20 & 9.36 & 0.43    \\
        \hline
        \multirow{4}{*}{PINAT~\cite{lu2023pinat}} &\multirow{4}{*}{Transformer}
        & MSE & 146.60  & 8.62  & 0.62   \\
       & & HR  & 8.44 & 29.32  & 0.67                   \\
       & & ListMLE  & 21.78 & 23.56 & \textbf{0.68}                    \\
       &  & WARP  & \textbf{3.78} & \textbf{38.71}   & 0.65                    \\
        \thickhline
\end{tabular}
}
\label{table:ap and pinat}
\end{table}

%% file: figs/search_res.tex
\begin{figure*}[ht]
     \begin{minipage}[b]{0.32\textwidth}
        \centering
        \includegraphics[width=\textwidth]{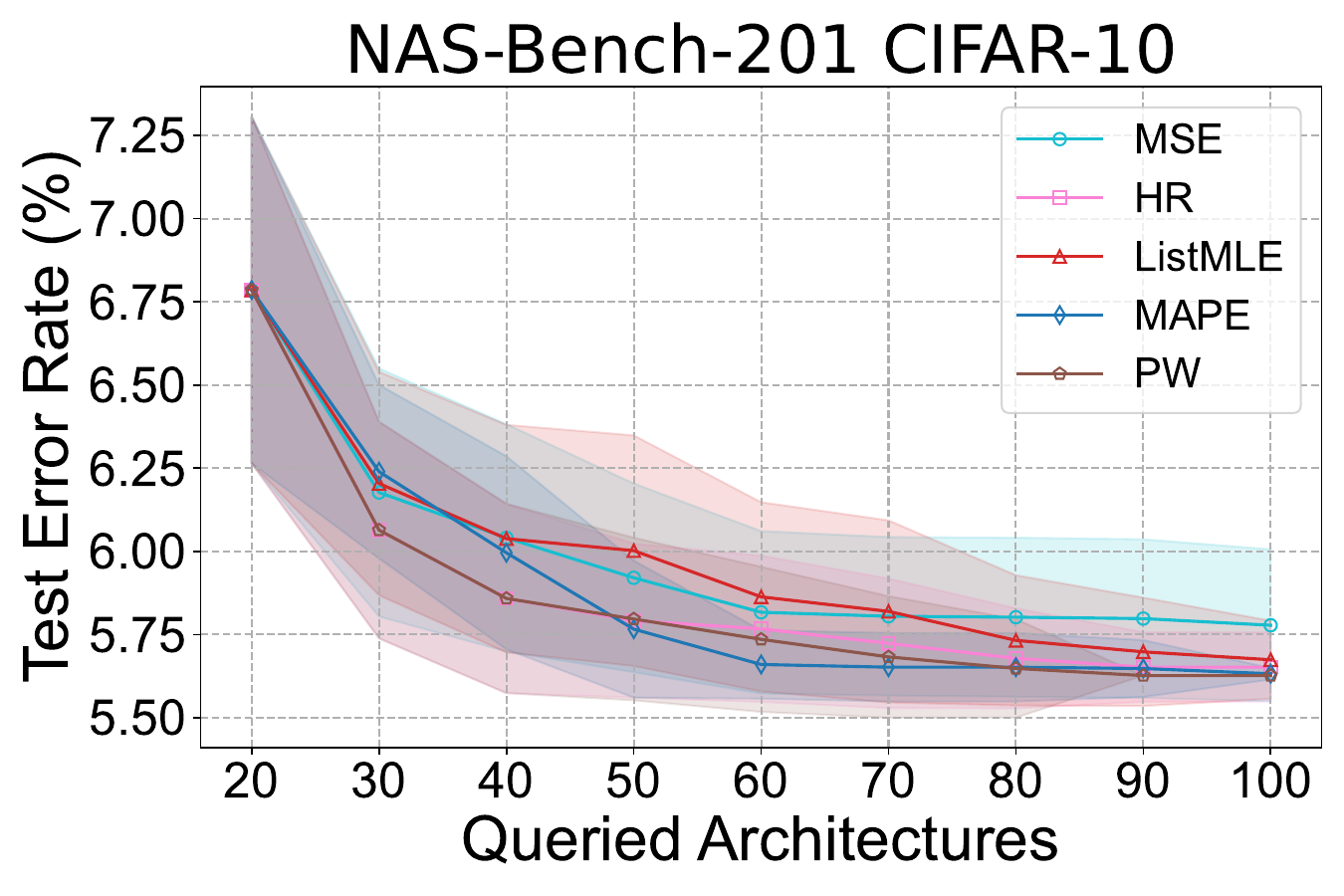}
    \end{minipage}  
     \begin{minipage}[b]{0.32\textwidth}
        \centering
        \includegraphics[width=\textwidth]{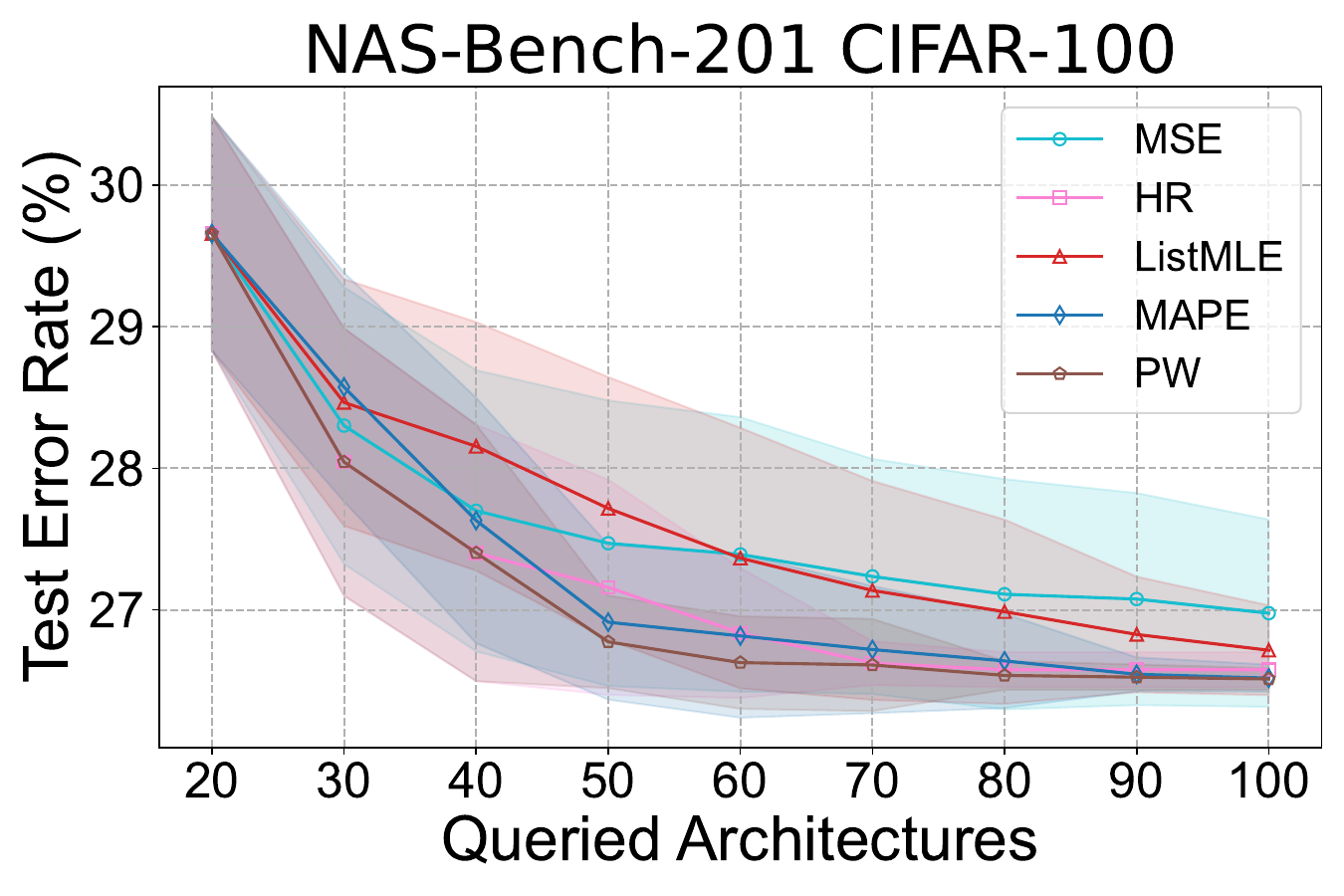}
    \end{minipage}  
    \begin{minipage}[b]{0.32\textwidth}
        \centering
        \includegraphics[width=\textwidth]{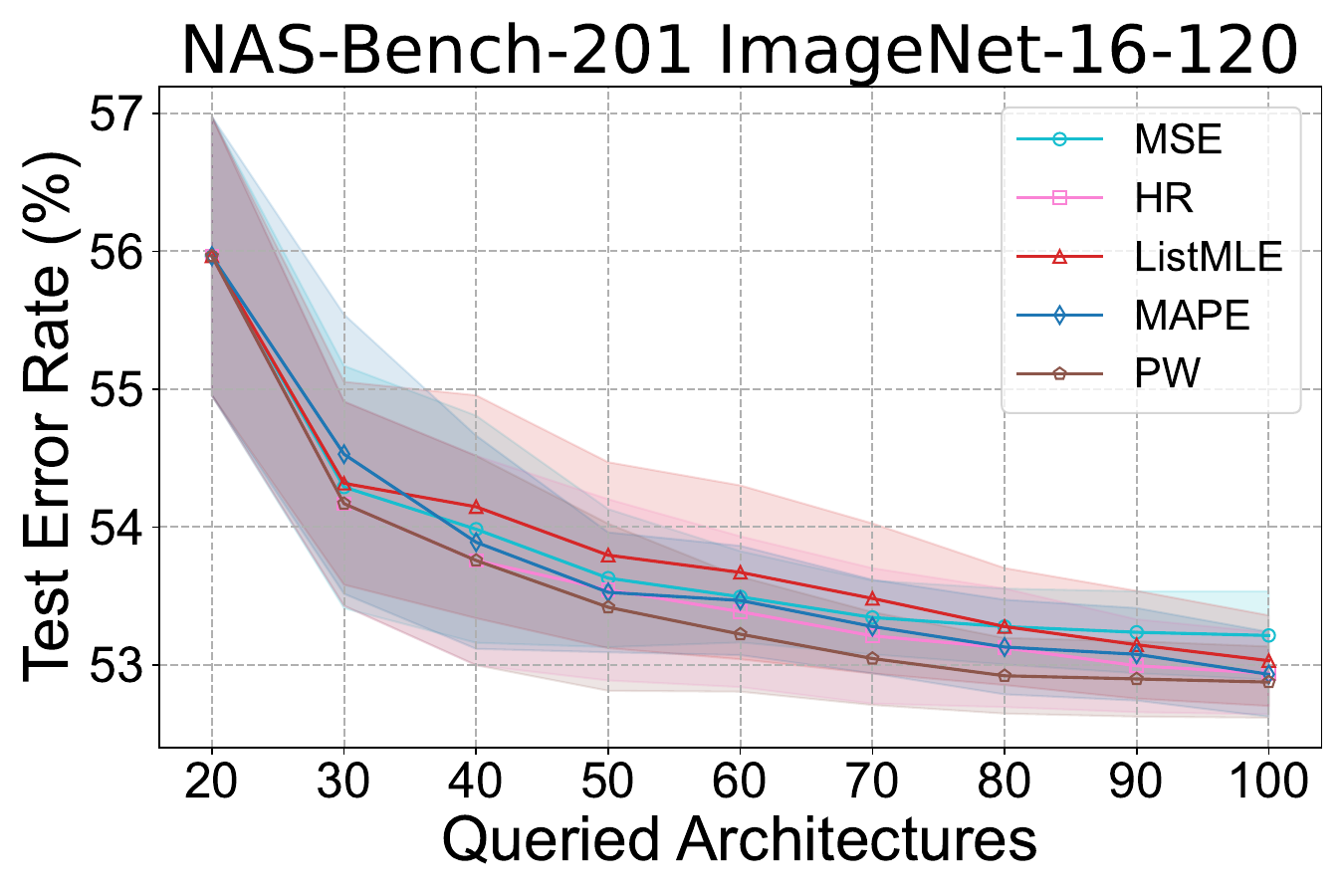}
    \end{minipage}

    \vspace{-0.5em}
     \caption{Comparisons between architectures searched by different types of loss functions on NAS-Bench-201.}
    \label{fig:search res 201}
\end{figure*}

%% file: tabs/search_201.tex
\begin{table}[t]
 \centering
   \renewcommand{\arraystretch}{1.12}
\caption{Searching results on NAS-Bench-201 with a query budget of 100. \textbf{Bold} indicates the best.}
 \resizebox{1\columnwidth}{!}{
        \begin{tabular}{c|c|c|c|c}
        \thickhline
        Predictor  & Loss  & CIFAR-10 & CIFAR-100 & ImageNet-16-120   \\
        \hline
        NASBOT~\cite{kandasamy2018neural} & MSE & 6.36  & 28.62 & 54.12      \\
        ReNAS~\cite{xu2021renas} & LR  & 6.01  & 27.88 & 54.03                      \\
        NPENAS~\cite{wei2022npenas} & MSE  & 5.69 & 26.54 & 53.52              \\
        \hline
        PWLNAS (ours) & PW & \textbf{5.63} & \textbf{26.51} & \textbf{52.88}              \\
        \hline
         Global Best & -  & 5.63  & 26.49 & 52.69  \\
        \thickhline
\end{tabular}
}
\label{table:search 201}
\end{table}

%% file: tabs/search_101.tex
\begin{table}[t]
 \centering
   \renewcommand{\arraystretch}{1.12}
\caption{Searching results on NAS-Bench-101 with a query budget of 150. \textbf{Bold} indicates the best.}
\vspace{-0.5em}
 \resizebox{1\columnwidth}{!}{
        \begin{tabular}{c|c|c|c}
        \thickhline
        Predictor  &Loss  & Strategy & Test Err.(\%)   \\
        \hline
        BANANAS~\cite{white2021bananas} & MAPE  &BO  & 5.92                       \\
        WeakNAS~\cite{wu2021stronger} & MSE & Random & 5.90   \\
        CATE~\cite{yan2021cate} & EW          & Reinforce  & 5.88                 \\  
        FlowerFormer~\cite{hwang2024flowerformer} & HR     &  Evolution       &    5.86           \\
        NPENAS~\cite{wei2022npenas} & MSE & Evolution      & 5.85                       \\
        \hline
        \multirow{5}{*}{PWLNAS (ours)}
        & MSE & \multirow{5}{*}{Evolution}  & 6.06     \\
        & HR &  & 5.83                        \\
        & ListMLE &  & 5.84                        \\
        & WARP  & & 5.86                         \\
         & PW &  & \textbf{5.80}                        \\
         \hline
         Global Best & -    & -  & 5.68  \\
        \thickhline
\end{tabular}
}
\label{table:search 101}
\end{table}
\vspace{-1em}

%% file: tabs/search_micro.tex
\begin{table}[t]
    \centering
    \renewcommand{\arraystretch}{1.12}
    \caption{Searching results on TransNAS-Bench-101 Micro with a query budget of 50. \textbf{Bold} indicates the best.}
       \resizebox{1\columnwidth}{!}{
    \begin{tabular}{c|c|cccc}
    \thickhline
   \multicolumn{2}{c|}{Tasks}  & Cls.O. & Cls.S. & Auto. & Jigsaw  \\
    \cline{1-6}
    \rule{0pt}{8pt}
  Predictor & Loss  & Acc.$^{\uparrow}$ & Acc.$^{\uparrow}$ & $\operatorname{SSIM}^{\uparrow}$  & Acc.$^{\uparrow}$  \\
    \hline 
 Arch-Graph~\cite{huang2022arch} & BCE & 45.48 & 54.70 & 56.52 & 94.66  \\
 WeakNAS~\cite{wu2021stronger} & MSE & 45.66 & 54.72 & 56.77 & 94.63 \\
   \hline
 \multirow{5}{*}{PWLNAS (ours)}  & MSE & 45.36 & 54.66 & 55.59 & 94.76   \\
   & HR  & 45.57 & 54.67 & 56.12  & 94.61 \\
   & ListMLE & 45.53 & 54.66 & 55.95  & 94.63  \\
   & WARP & 45.46 & 54.76 & 56.87  & 94.62  \\
   & EW & 45.51 & 54.71  & 55.88 & 94.67 \\
   &PW & \textbf{45.73} & \textbf{54.83} & \textbf{56.93}  & \textbf{94.88}  \\
    \hline
 \multicolumn{2}{c|}{Global Best} & 46.32 & 54.94 & 57.72  & 95.37 \\
\bottomrule
\end{tabular} 
}
    \label{table: micro search}
\end{table}

%% file: tabs/darts_cf10.tex
\begin{table}[t]
  \centering
  \renewcommand{\arraystretch}{1.12}
  \caption{Searching results on the CIFAR-10 dataset within DARTS. The search cost is evaluated with GPU Days. \textbf{Bold} indicates the best.}
  \resizebox{1\columnwidth}{!}{
    \begin{tabular}{c| c |c| c| c}
        \thickhline
          Predictor & Loss & Test Err.($\%$) & Params(M) & Cost    \\
        \hline
         PNAS~\cite{liu2018progressive}  & MSE  & 3.34$\pm$0.09 & \textbf{3.2} & 225  \\
         TNASP~\cite{lu2021tnasp} &   MSE   & 2.57$\pm$0.04 & 3.6 & 0.3  \\
         CDP~\cite{liu2022bridge} & MSE  & 2.63$\pm$0.08 & 3.3 & \textbf{0.1}    \\
         NPENAS~\cite{wei2022npenas} & MSE & 2.54$\pm$0.10 & 3.5 & 1.8   \\
         PINAT~\cite{lu2023pinat} & MSE & 2.54$\pm$0.08 & 3.6 & 0.3   \\
         GMAENAS~\cite{jing2022graph} & BPR  & 2.50$\pm$0.03 & 3.6 & 3.3 \\
         DCLP~\cite{zheng2024dclp} & ListMLE & 2.48$\pm$0.02 & 3.3  & 0.14   \\         
        \hline
        \multirow{5}{*}{PWLNAS (ours) } & MSE   
        & 2.74$\pm$0.04 & 4.4 &\multirow{5}{*}{0.2}  \\
         & HR  &  2.67$\pm$0.09 & 3.9 &  \\
         & ListMLE  & 2.62$\pm$0.07 & 3.8 &   \\
        & MAPE & 2.53$\pm$0.06 & 4.5 &   \\
        & PW & \textbf{2.47$\pm$0.05} & 3.6 &   \\

    \thickhline
    \end{tabular}
    }
    \label{table:darts cf10}
\end{table}

%% file: sec/5_suggestions.tex
\section{Discussions and Suggestions}
\label{sec:suggestions}
In Section~\ref{sec: evaluation}, we evaluate loss functions of performance predictors under various settings and observe similar trends in most cases. The key findings include: (1) Weighted loss functions identify good architectures better with sufficient training data while ranking ones tend to perform better with extremely few training data; (2) Simple predictors (MLP-based) work better with ranking loss functions and advanced predictors (GCN/Transformer-based) benefit more from weighted ones in distinguishing promising architectures. (3) Specific categories of loss functions can be combined to enhance predictor-based NAS. Our results can recommend suitable loss functions for predictor-based NAS in new tasks majorly according to their performance in top-ranking metrics like N@K and Precision@T. For instance, if a predictor-based method aims to explore a search space that resembles NAS-Bench-201 using a GCN-based predictor, we suggest employing HR loss in the early iterations to warm up and then using MAPE loss to train the predictor.

Meanwhile, we also offer several suggestions to improve predictor-based NAS according to the findings revealed in this paper. First, \textbf{combine different types of loss functions in a more flexible way}. Although the proposed PW loss is simple and effective, the value of the threshold to use different losses is fixed and relies on human experience. A promising direction is increasing the intensity of weight for loss functions as the training portion grows. In the beginning, the weight can be set to 0, which equals to ranking loss. In this way, the weighted loss is expected to better distinguish good architectures on different constraints of training portions. Second, \textbf{design loss functions that directly optimize top-K metrics}. Despite the effectiveness of existing weighted loss functions, they optimize top-K metrics by assigning weights indirectly. Since several measure-specific loss functions~\cite{wang2018lambdaloss,jagerman2022optimizing} have achieved impressive results in other top-ranking problems, involving the top-K metrics such as Precision@T during optimization is also feasible for the loss function in predictors, which can further enhance the predictor-based NAS. Additionally, \textbf{designing advanced sampling strategies to select representative architectures for training the predictor}. We observe that more training samples are not certain to benefit the top-ranking ability of predictors even with a weighted loss function, which could stem from that randomly selected training samples fail to represent the distribution of the diverse search space. An effective sampling method can greatly improve the efficiency of predictor-based NAS.

%% file: sec/6_conclusion.tex
\section{Conclusion}
\label{conclusion}
In this paper, we provide the first comprehensive study for eight loss functions in performance predictors, including regression, ranking, and weighted loss functions. We compare their performance using various assessment metrics on 13 NAS tasks across five search spaces, including cell-based and skeleton-based ones. Meanwhile, we leverage the advantage of specific categories of loss functions and propose a new PWLNAS method that employs a piecewise loss function. PWLNAS achieves better results than using any single loss function and outperforms competitive predictor-based NAS methods. We hope this paper can motivate the design of new loss functions to enhance predictor-based NAS in further research.